\documentclass{article}

\usepackage{PRIMEarxiv}

\usepackage[utf8]{inputenc} 
\usepackage[T1]{fontenc}    
\usepackage{hyperref}       
\usepackage{url}            
\usepackage{booktabs}       
\usepackage{amsfonts}       
\usepackage{nicefrac}       
\usepackage{microtype}      
\usepackage{lipsum}
\usepackage{fancyhdr}       
\usepackage{graphicx}       
\graphicspath{{media/}}     

\usepackage{graphicx}
\usepackage{caption}
\usepackage{subcaption}
\usepackage{fvextra}
\usepackage{amsfonts}
\usepackage{multirow}
\usepackage[table,xcdraw]{xcolor}
\usepackage{epstopdf}
\usepackage{fancyvrb}
\usepackage{xcolor}
\usepackage{amsmath}
\usepackage{natbib}

\title{Neurosymbolic Inference on Foundation Models for Remote Sensing Text-to-image Retrieval with Complex Queries
\thanks{\textit{\textbf{Accepted for publication in ACM Transactions on Spatial Algorithms and Systems (TSAS).}}} 
}

\author{
  Emanuele Mezzi \\
  Vrije Universiteit Amsterdam \\
  The Netherlands \\
  \texttt{e.mezzi@vu.nl} \\
   \And
  Gertjan Burghouts \\
  TNO \\
  The Netherlands \\
  \texttt{gertjan.burghouts@tno.nl} \\
   \And
  Maarten Kruithof \\
  TNO \\
   The Netherlands \\
  \texttt{maarten.kruithof@tno.nl} \\
}

\definecolor{myred}{RGB}{186,22,22}
\definecolor{mygreen}{RGB}{29,164,29}
\definecolor{myorange}{RGB}{247,167,27}
\definecolor{mypurple}{RGB}{156,27,230}

\begin{document}

\newcommand{\manu}[1]{{\color{black} #1}}

\maketitle

\begin{abstract}
Text-to-image retrieval in remote sensing (RS) has advanced rapidly with the rise of large vision-language models (LVLMs) tailored for aerial and satellite imagery, culminating in remote sensing large vision-language models (RS-LVLMS). However, limited explainability and poor handling of complex spatial relations remain key challenges for real-world use. To address these issues, we introduce RUNE (Reasoning Using Neurosymbolic Entities), an approach that combines Large Language Models (LLMs) with neurosymbolic AI to retrieve images by reasoning over the compatibility between detected entities and First-Order Logic (FOL) expressions derived from text queries. Unlike RS-LVLMs that rely on implicit joint embeddings, RUNE performs explicit reasoning, enhancing performance and interpretability. For scalability, we propose a logic decomposition strategy that operates on conditioned subsets of detected entities, guaranteeing shorter execution time compared to neural approaches. Rather than using foundation models for end-to-end retrieval, we leverage them only to generate FOL expressions, delegating reasoning to a neurosymbolic inference module. For evaluation we repurpose the DOTA dataset, originally designed for object detection, by augmenting it with more complex queries than in existing benchmarks. We show the LLM's effectiveness in text-to-logic translation and compare RUNE with state-of-the-art RS-LVLMs, demonstrating superior performance. We introduce two metrics, Retrieval Robustness to Query Complexity (RRQC) and Retrieval Robustness to Image Uncertainty (RRIU), which evaluate performance relative to query complexity and image uncertainty. RUNE outperforms joint-embedding models in complex RS retrieval tasks, offering gains in performance, robustness, and explainability. We show RUNE's potential for real-world RS applications through a use case on post-flood satellite image retrieval.
\end{abstract}


\section{Introduction}

Given their impressive capabilities, foundation models and large vision-language models (LVLMs) are seeing rapidly expanding adoption, with applications spanning nearly every area of scientific research and practical implementation \cite{lisurvey,van2024large,wang2024leveraging,li2024enhancing}. In recent years, LVLMs have been successfully applied to remote sensing (RS) \cite{tao2025advancements,li2024vision}, where GeoAI foundation models can handle a variety of tasks — notably text-to-image retrieval, which enables automated extraction of actionable information from RS data. Models such as RemoteCLIP \cite{liu2024remoteclip}, GeoRSCLIP \cite{zhang2024rs5m}, and RS-LLaVA \cite{bazi2024rs} have shown promising retrieval performance, achieving high recall and ensuring good coverage of relevant images given a textual query. These results highlight that, even in RS, general-purpose LVLMs can deliver strong performance when fine-tuned on data tailored to the task.

However, RS-LVLMs face several key limitations. First, they require large amounts of training data, which is often unavailable in the target domain, motivating the need for models that can generalize without domain-specific training. Second, existing LVLMs rely on joint image-text embeddings, where both the query and image are projected into a shared embedding space and their similarity is computed. While effective, this approach encodes the detailed semantics of the query implicitly, limiting both representational power and interpretability. Explainability is critical for real-world applications, where human-understandable reasoning is often required. More importantly, we show that LVLMs struggle with semantically complex queries — especially those involving multiple objects and relationships.

Previous evaluations have largely focused on recall (R@k), which provides useful insights into coverage and false negatives (FNs) but fail to capture models’ ability to limit false positives (FPs), an important consideration in operational settings, where manual intervention on every alert is impractical. To address this, we conduct a thorough evaluation of RS-LVLMs on a dataset characterised by queries with complex spatial relations, assessing both recall and precision (P@k), and analyzing robustness across increasing levels of query complexity. Our results reveal that current LVLMs lack precision, robustness, and interpretability.

To overcome these challenges, we propose RUNE (Reasoning Using Neurosymbolic Entities), a novel text-to-image retrieval approach that integrates foundation models with logical reasoning. Our method builds on the observation that language queries such as “five cars aligned and a ship at the left of a harbor” can be effectively expressed in First-Order Logic (FOL). RUNE uses FOL-based reasoning for retrieval. Rather than relying on foundation models to perform retrieval end-to-end, we harness them for a task they excel at: transforming text into FOL while using object detectors to identify objects in images. The reasoning itself is handled by a neurosymbolic inference module, which accounts for the probabilistic nature of entity detection and the relationships between them. As a result of this neurosymbolic inference approach, RUNE consists of a zero-shot learning approach for text-to-image retrieval as it does not require any training to perform it. Figure \ref{fig:methodology_pipeline} provides an overview of our approach.

Our contributions are as follows:

\begin{itemize}
\item A novel text-to-image retrieval method called RUNE, which combines text-to-logic translation with neurosymbolic reasoning, enabling explicit representation of entities and their relationships, and achieving improved performance, robustness, and explainability.
\item A new form of in-context learning for RS, to translate natural language queries into FOL, named spatial logic in-context learning for remote sensing (SLCLRS), enhancing LLM reasoning and interpretation.
\item An efficient neurosymbolic inference procedure using logical decomposition on conditioned subsets of entities, ensuring RUNE’s scalability.
\item \manu{Two novel metrics, Retrieval Robustness to Query Complexity (RRQC) and Retrieval Robustness to Image Uncertainty (RRIU), designed to quantitatively assess the robustness of RUNE and baseline approaches as query complexity and image uncertainty increase, respectively.}
\item \manu{A use case demonstrating the effectiveness of RUNE for retrieving post flood satellite imagery, leveraging the FloodNet dataset \cite{rahnemoonfar2021floodnet}, illustrating how RUNE can be used in disaster-related scenarios.}
\item An extension of the DOTA \cite{xia2018dota} dataset for text-image retrieval, augmenting it with queries of varying complexity — including diverse entities, classes, and relationships — going beyond the capabilities of current benchmark datasets.
\end{itemize}

Our manuscript is organized as follows: Section \ref{sec:background} summarizes the background regarding LLMs, LVLMs, and Neurosymbolic AI. Section \ref{sec:related_work} discusses the related work in the field of RS text-image retrieval. Section \ref{sec:methodology} describes our proposed methodology in all its components. Section \ref{sec:experiments} presents the experimental set-up. Section \ref{section:comparison_with_rs_lvms} presents the experiment results. Section \ref{sec:discussion} discusses the results and highlights interesting aspects, and Section \ref{sec:conclusions} draws the conclusions. 

\section{Background}
\label{sec:background}

\subsection{Large Language Models (LLMs)}
LLMs \citep{kasneci2023chatgpt,zhao2023survey} are sophisticated neural networks distinguished by their extensive parameter sets and remarkable learning abilities. They facilitate user interaction through prompt engineering \citep{zhou2022large,white2023prompt,clavie2023large}. This crucial technique involves carefully designing input prompts to steer the model toward specific tasks or desired outputs and forms a cornerstone of many current evaluation methodologies. Users can engage with LLMs via various modalities, including structured question-answering \citep{jansson2021online} and open-ended dialogues, leveraging the models' generative capabilities for tasks such as information retrieval and conversation. The essence of LLMs can be found in three components which consist of the transformer architecture \citep{vaswani2017attention}, the capacity to perform in-context learning, and the possibility to refine the quality of the output generated through Reinforcement Learning from Human Feedback (RLHF). These three component enables LLMs to represent a significant advancement in NLP, with broad implications across numerous application areas.

A key characteristic of LLMs is their capacity for in-context learning \citep{brown2020language}, which allows them to adapt to specific tasks based on contextual input, generating outputs conditioned on this information. This capability facilitates the generation of coherent and contextually relevant responses, making LLMs particularly well-suited for dialogue and human interaction. Furthermore, beyond in-context learning, which requires no additional training, LLMs can be adapted for specific tasks through RLHF \citep{ziegler2019fine,christiano2017deep}. This method involves fine-tuning models using human-annotated responses as a reward signal, helping to align model outputs with human preferences. The primary task performed by LLMs is predicting the subsequent token $y$ given a sequence of content $X$. Consequently, training involves maximizing the likelihood of a target sequence conditioned on the preceding context tokens, expressed as $P(y | X) = P(y | x_1, x_2, ..., x_{t-1})$, where \(x_1, x_2, ..., x_{t-1}\) represent the tokens preceding the current prediction point $t$.

\subsection{Large Visual Language Models (LVLMs)}
LVLMs represent a significant advancement, integrating language understanding with visual perception to enable the processing and reasoning of both text and images \cite{liu2025visual}. The development of LVLMs focuses on improving the synergy between textual and visual modalities \cite{zhang2024vision}. Over time, architectural strategies have evolved from training models from the ground up, as seen in CLIP \cite{radford2021learning,yao2022detclip}, to leveraging pretrained LLMs \cite{grattafiori2024llama3herdmodels} as core components to enhance the correlation between vision and language, thereby improving the comprehension of image content. Fundamentally, whether models are trained from scratch or built on pretrained LLMs, they incorporate the same key components. Vision encoders play a crucial role in converting visual inputs into embedding features that can be aligned with LLM embeddings for tasks such as text or image generation \cite{fini2024multimodal}. These encoders are designed to extract rich and informative representations from image or video data, facilitating integration with language features \cite{maaz2024videogpt+, zhao2024videoprism}. Typically pretrained on large-scale image-text or video-text datasets, vision encoders capture semantic relationships between visual and textual modalities. Notable examples include CLIP \cite{radford2021learning}, which aligns image and text embeddings using contrastive learning, and BLIP \cite{li2022blip}, which employs a bootstrapping-based training approach. Contemporary state-of-the-art LVLMs often utilize pretrained vision encoders, which not only provide high-quality visual embeddings \cite{wang2024qwen2, liu2023visualinstructiontuning} but also enhance adaptability \cite{zamir2018taskonomydisentanglingtasktransfer} by leveraging prior visual knowledge acquired from extensive datasets like ImageNet \cite{ImageNet}, thus outperforming alternatives based on random initialization \cite{holmberg2020self}.

Text encoders are responsible for mapping text into an embedding space. Architectures like CLIP \cite{radford2021learning}, OSCAR \cite{li2020oscar}, and ALIGN \cite{jia2021scaling} train both image and text encoders from scratch, using contrastive learning to align visual and textual embeddings within a shared joint embedding space, enabling effective cross-modal associations. In contrast, architectures such as LLaVA \cite{liu2023visualinstructiontuning} forgo separate text encoders, instead relying on the language understanding capabilities of LLMs like LLaMA \cite{touvron2023llama2openfoundation} and Vicuna \cite{peng2023instructiontuninggpt4}. In these models, visual information is integrated through mechanisms like projection layers or cross-attention modules \cite{tan2019lxmert}. This trend highlights a preference for using LLMs as central components in multimodal systems, allowing for more versatile and powerful reasoning and generation across different modalities. Text decoders, particularly in transformer-based architectures like GPT, generate human-readable text from internal representations. While training LVLMs from scratch necessitates a distinct text decoder, employing LLMs as the main component allows for the utilization of the LLMs' original decoders. Examples of decoder architectures used in multimodal pretraining include VisualBERT and VilBERT \cite{lu2019vilbertpretrainingtaskagnosticvisiolinguistic, li2019visualbertsimpleperformantbaseline}.

Cross-attention mechanisms facilitate the interaction between textual and visual modalities by enabling mutual influence between tokens from different modalities \cite{chen2021crossvit}. However, it is important to note that not all VLMs utilize cross-attention mechanisms. While models such as LLaVA \cite{liu2023visualinstructiontuning} and VL-T5 \cite{cho2021unifying} incorporate it, CLIP \cite{radford2021learning} does not rely on it.

\subsection{Neurosymbolic AI}

Neurosymbolic AI is an increasingly important area of artificial intelligence that synergistically combines the pattern recognition and data-handling capabilities of neural networks with the structured reasoning offered by symbolic systems. This integrated approach allows for the fusion of knowledge learned from data with abstract, rule-based representations, thereby enhancing AI systems' abilities in tasks demanding reasoning, planning, and conceptual understanding \cite{ahmed2022pylon,anderson2020neurosymbolic,li2023scallop,manhaeve2021neural,mitchener2022detect}. By bridging symbolic and sub-symbolic paradigms, neurosymbolic systems leverage their respective strengths, yielding solutions that are both robust and cognitively plausible \cite{d2020neurosymbolic,marcus2020next,wu2023verix}.

Sheth et al. \cite{sheth2023neurosymbolic} analyze the field through two perspectives: an algorithmic one, focusing on the computational attributes of neurosymbolic models like abstraction, analogy, and long-term decision-making; and an application-oriented one, highlighting how these systems improve practical aspects such as interpretability, explainability, and safety in real-world applications. Algorithmic parallels can be drawn to Daniel Kahneman’s dual-system theory of cognition \cite{kahneman2011thinking}, where System 1 represents fast, intuitive processing and System 2 embodies deliberate, logical thought. In this analogy, deep learning models correspond to System 1, efficiently processing large volumes of raw data and excelling in tasks like image recognition or language modeling. Symbolic systems, akin to System 2, while potentially lacking scalability with extensive datasets, are invaluable for structured reasoning and formal knowledge manipulation. The fundamental concept of neurosymbolic AI is to merge these systems, integrating the perceptual strengths of neural networks with the logical rigour of symbolic reasoning.

From a practical standpoint, the integration of these two forms of intelligence addresses critical requirements for the safe and trustworthy deployment of AI \cite{kosasih2024review}. Neural networks, despite their power, often operate as black boxes and are vulnerable to failures arising from biased data \cite{tjuatja2024llms}, adversarial attacks \cite{ren2020adversarial}, and distribution shifts \cite{carlini2017towards}. While Explainable AI (XAI) techniques aim to interpret neural models, their outputs frequently remain inaccessible to non-expert users. To develop AI systems that are transparent and reliable, incorporating structured background knowledge from the outset, rather than relying solely on post-hoc explanations, becomes crucial. Symbolic representations offer a means to directly embed domain rules and constraints into the system \cite{von2021informed}, enhancing both interpretability and robustness. This is particularly vital in high-stakes environments where AI decisions must be clear, justifiable, and secure \cite{sheth2022process}.

\section{Related work}
\label{sec:related_work}

The emergence of foundation models, demonstrating remarkable cross-modal performance, has led the RS community to increasingly explore LVLMs for tasks such as text-image retrieval. Recent advancements in RS text-image retrieval have focused mostly on LVLMs, especially to develop representative joint embedding spaces, with a strong emphasis on multi-level or multi-scale image representations \cite{yuan2021lightweight,yuan2022exploring,yuan2022remote,pan2023reducing}. These methodologies aim to better capture the intricate visual semantics inherent in RS imagery and align them with textual descriptions. Liu et al. \cite{liu2024remote} introduced RemoteCLIP, the first vision-language foundation model specifically tailored for RS. Their approach emphasizes learning robust and semantically rich visual features alongside aligned text embeddings, facilitating seamless transfer to downstream applications. Building on this trend, Zhang et al. \cite{zhang2024rs5m} proposed a framework incorporating a Domain-pretrained Vision-Language Model (DVLM) to bridge the gap between general-purpose vision-language models (GVLMs) and RS-specific tasks. They also introduced RS5M, the first large-scale image-text paired dataset in RS, comprising 5 million images with English annotations.

Mall et al. \cite{mall2023remote} presented GRAFT (Ground Remote Alignment for Training), a novel technique for training vision-language models for RS-tasks without relying on textual annotations. Their method uses co-located ground-level internet imagery as an intermediary between satellite images and natural language, enabling training without labeled data. These LVLM-based methods have shown impressive results on zero-shot and open-vocabulary tasks such as classification, retrieval, segmentation, and visual question answering for satellite imagery. Silva et al. \cite{silva2024large} proposed RS-CapRet, which utilizes a large decoder-based language model along with image encoders adapted to RS imagery through contrastive language-image pretraining. To connect the encoder and decoder, they train lightweight linear projection layers using diverse RS captioning datasets while keeping the remaining model parameters frozen. RS-CapRet achieves state-of-the-art or highly competitive performance in image captioning and text-based image retrieval tasks.

Despite these promising advancements, current RS-LVLMs still encounter significant challenges, particularly concerning representational power to capture complex queries, explainability and effective evaluation in complex retrieval scenarios—especially those requiring an understanding of inter-object relationships. To address the semantic gap between textual queries and image content, Mi et al. \cite{mi2024knowledge} introduced an approach that integrates external knowledge graphs to enrich the textual input, thereby improving alignment with visual content. However, while this method enhances textual scope, it does not fully overcome limitations in reasoning or interpretability.

Previous approaches solved the problem of text-image retrieval by training and using models based on the generation of a joint text-image embedding, where both the search query and the image are projected and the similarity between them is calculated. For improved performance, robustness and explainability, we approach the problem differently, by logical reasoning using neurosymbolic AI with foundation models as components.

\section{Methodology}
\label{sec:methodology}

\begin{figure}[h]
    \centering
    \includegraphics[width=\linewidth]{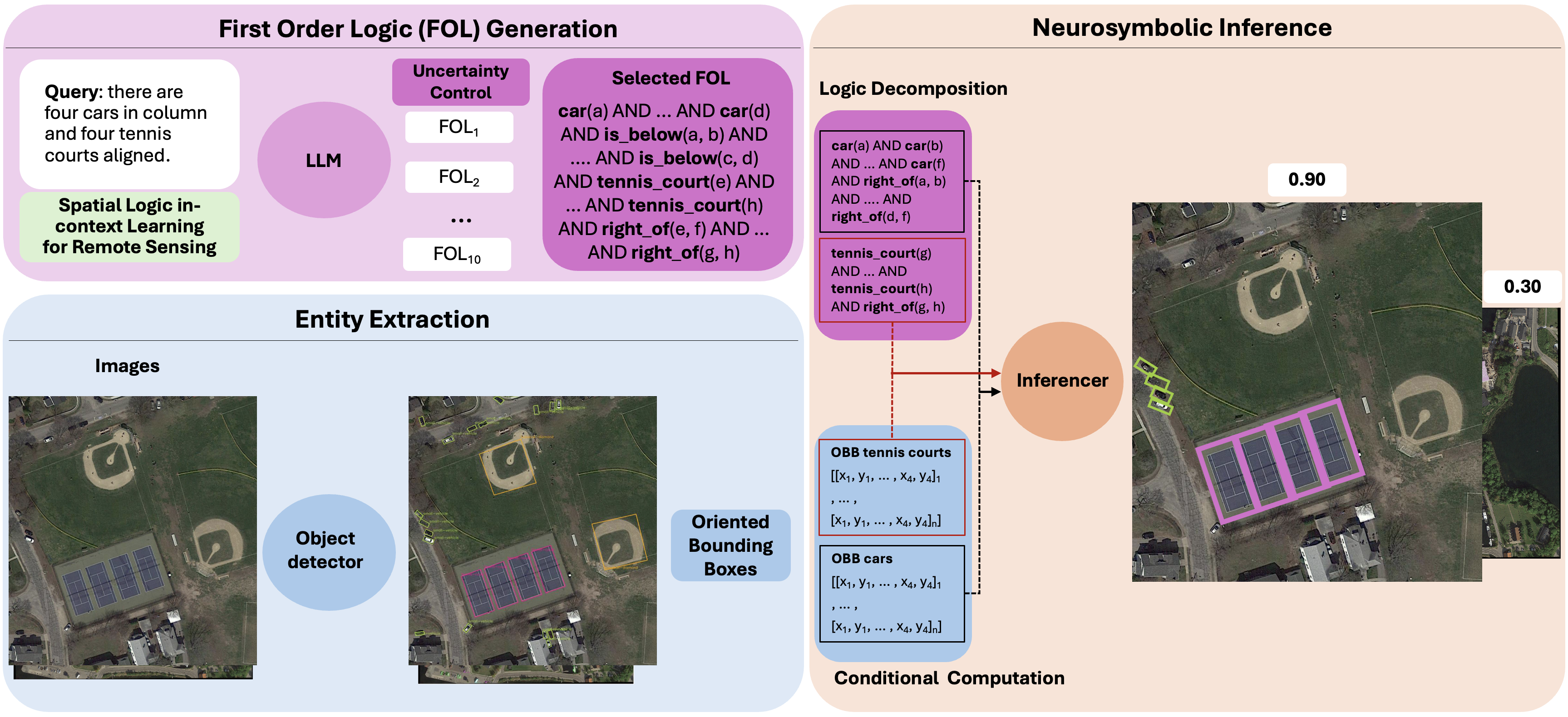}
    \caption{Our method performs text-image retrieval by logical reasoning, combining foundation models for text-to-logic translation and entity extraction, with neurosymbolic inference for entities and relations between them.}
    \label{fig:methodology_pipeline}
\end{figure}

Our text-image retrieval method outlined in Figure \ref{fig:methodology_pipeline} integrates three building blocks: 

\begin{itemize}
    \item Text-to-logic transformation which converts the natural language query from a user to the corresponding FOL expression about the involved entities and their relationships. The text-to-logic transformation is tailored to RS and the understanding of the possible relations between entities in language queries through in-context learning. Moreover, we handle the lack of determinism in LLMs by applying an uncertainty control based on multi-sampling.
    \item Entity extraction, using object detection to predict a location (an oriented bounding box) and associated probability for each candidate entity in an image.
    \item Neurosymbolic reasoning on each image, driven by the FOL expression, operating on the detected objects and their relations, to infer a probability for an image that it matches the FOL and thereby the query that the user is interested in. Our inference is computationally tractable, reducing the intractable amount of hypotheses by conditional computations, and by decomposing the logic and entities.
\end{itemize}

\subsection{First-Order Logic for Remote Sensing Queries}
\label{subsec:first_order_logic_rs}

The conversion of the queries from text to logic is performed by combining LLMs operations and rule-based post-processing. To perform this conversion, we introduce a novel in-context learning technique named \textbf{spatial logic in-context learning for remote sensing (SLCLRS)}, which combines the task of text-to-logic translation with entities specific to the RS context encountered by the LLM in the queries. The prompt is composed of the following sections: 

\begin{itemize}
    \item \textbf{Role section}: we assign the LLM a specific role, that consists of being an expert logician.
    \item \textbf{Context section}: we provide the LLM with the task context, specifying the entities, objects, relations, and logical predicates to be considered.
    \item \textbf{Text-to-logic section}: we provide the LLM with instructions regarding the text-to-logic conversion, dividing the instructions into two steps to maximize the understanding by the LLM. 
    \item \textbf{Few-shot section}: we provide five few-shot examples to implement in-context learning. For each example we provide the language query and the corresponding FOL expression that must be produced. 
    \item \textbf{Output section}: we specify that the output must follow the JSON format, characterized by two keys: variables and translations. The variables key contains the associations between object types and their corresponding variables, while translations store the natural language query alongside the resulting logical expression.
\end{itemize}

Of these five sections, the Role, Context, and Few-shot section, implement the SLCLRS technique to help the LLM focus on specific entities and predicates of the RS scenario, while ensuring a more precise translation. \manu{The list of entities which the queries can contain is presented in Table \ref{tab:object_predicates}. In case the LLM receives a queries with entities that cannot be processed by the system it communicates this to the user that thus can modify the query}. These sections collectively convey the role the LLM must embody, the RS context, and examples to guide accurate translation. Figure \ref{fig:query-to-logic} shows the prompt used to perform the conversion from natural language query to FOL expression. For each natural language query composed of $n$ words, and $m$ connections, each of which corresponds to a specific relation, the text-to-logic converter outputs the FOL expression, respecting the following format:

\begin{align}
    \label{eq:logic_transform}
    LLM(text\_query) = obj_{1}(A) \: AND ... \: AND \: obj_{n}(Z) \: AND \: rel_{1}(A, B) \: ... \: rel_{m}(J, Z)
\end{align}
where
\begin{itemize}
    \item $A \: ... \: Z$: each different object in the natural language query is represented by a different letter of the alphabet. 
    \item $obj_{i}$: represents the category of the object that the LLM can highlight in the sentence. 
    \item $rel_{i}$: represents the relationship type that connects two objects. 
\end{itemize} 

Here follows an example of transformation from language query to the corresponding logic format. Specifically, we provide an example composed of relationships that involve only two objects: 

\begin{gather*}
    \label{eq:example_1}
    \color{blue}\textbf{query\_language} \color{black} = \text{two ships close to each other.}\\
    \color{red}\textbf{query\_FOL} \color{black} = ship(A) \: AND \: ship(B) \: AND \: is\_close(A, B)
\end{gather*}

If the relationship is characterized by connecting more than two objects, it is decomposed into its atomic components, as shown by the following example: 

\begin{gather*}
    \label{eq:example_2}
    \color{blue}\textbf{query\_language} \color{black} = \text{three ships aligned.}\\
    \color{red}\textbf{query\_FOL} \color{black} = ship(A) \: AND \: ship(B) \: AND \: ship(C) \: AND \\ left\_of(A, B) \: AND \: left\_of(B, C)
\end{gather*}

In this example, the language query contains the relation \emph{aligned} which can be decomposed by a series of atomic relationships highlighting how each object is to the left or the right of the other object. 

\begin{figure}
\centering
\begin{minipage}[h]{\linewidth}
{\small
\begin{Verbatim}[commandchars=\\\{\},frame=single,framerule=1.2pt, breaklines=true,breakindent=0pt,breakautoindent=false,breaksymbolleft={}]
\textcolor{myred}{You are a \textbf{logician} and expert in \textbf{Remote Sensing (RS)}.}

\textcolor{mygreen}{You are working on queries that are used to retrieve images in a \textbf{remote sensing (RS)} scenario. I want you to transform queries in natural language to the corresponding first order logic (FOL) expression. Following, there is the list of the entities and the predicates that you can extract.}

\textcolor{mygreen}{\textbf{Objects}:

plane, ship, storage tank, baseball diamond, tennis court, basketball court, ground track field, harbor, bridge, truck, car, roundabout, soccer ball field and swimming pool.}

\textcolor{mygreen}{\textbf{Relations}:

is on, is close, left of, right of, is above, is below, is different, facing same, facing opposite, aligned, in column, isolated from, clustered.}

\textcolor{myorange}{\textbf{Steps to follow}:

    - \textbf{Step 1:} assign to each object in the query a variable, correspinding to an alphabet letter. Assign to each object a different letter. 
    - \textbf{Step 2:} parse the query into a logical expression using the variables above and the keywords.}

\textcolor{blue}{\textbf{Conversion examples:}

\textbf{Example 1} 
    - Query: two trucks close to each other
    - FOL expression: truck(a) and truck(b) and is_close(a, b)
...
\textbf{Example 5} 
    - Query: six ships in column and four cars aligned. 
    - FOL expression: ship(a) and ship(b) and ship(c) and ship(d) and ship(e) and ship(f)
}
\textcolor{mypurple}{Return the answer in the following \textbf{JSON} format: 
\{
    \textbf{variables}: \{"a": "category1", "b": "category2", "c": "category3"\}
    \textbf{translations}: [\{"query": query string, "expression": FOL expression\}]
\}
}
Convert to FOL the following query: \textbf{user query}.
\end{Verbatim}
}
\end{minipage}
\caption{Prompt to instruct the LLM to perform the conversion of queries from the natural language format to the equivalent FOL expression. The prompt sections are: \textcolor{myred}{Role} that assigns a role to the LLM, \textcolor{mygreen}{Context} that provides the LLM with context regarding the task to perform, \textcolor{myorange}{Text-to-logic} that provides the LLM with the instructions to perform the translation. \textcolor{blue}{Few-shot} section, provides the LLM with examples to correctly perform the translation. \textcolor{mypurple}{Format} section provides the LLM with instructions regarding the output JSON format. The \textcolor{myred}{Role}, \textcolor{mygreen}{Context}, and \textcolor{blue}{Few-shot} sections together provide the \textbf{spatial logic in-context learning for remote sensing}, which allows the LLM to know the entities and the predicates which are involved in the RS scenario.}
\label{fig:query-to-logic}
\end{figure}

Given the inherent non-determinism of LLMs \cite{xiong2024efficient}, and the fact that the LLM is part of a larger pipeline, where such non-determinism could affect downstream components \cite{mezzi2024risks}, we implement an uncertainty control mechanism to prevent its propagation. By implementing a multi-sampling technique, we prompt the LLM ten times with the same query, record all resulting FOL expressions, and ultimately select a single expression. To select only one FOL expression, we first calculate the embedding similarity between the language query and the FOL expressions. If multiple queries have the same cosine similarity, we select the FOL expression with the highest confidence output by the LLM among them. If the LLM generates multiple FOLs with the same confidence, we select the FOL expression with the lowest number of entities and predicates. This way, by adopting a conservative approach to the number of entities and predicates, we limit the possibility of introducing False Positive entities and predicates in the FOL expression.

\subsection{Entity Extraction for Remote Sensing Imagery}

From each image, we extract the objects that are part of the image. To accomplish object extraction, we perform oriented bounding box (OBB) detection, which identifies the location and type of objects within the image. We used OBB detection because, unlike horizontal bounding box (HBB) detection, it captures not only the location but also the orientation of objects, thereby reducing the amount of background information and conveying important information regarding its position and orientation \cite{wen2023comprehensive}, fundamental factors in establishing relations between objects. The use of OBB detection is fundamental for the elaboration of the query used to perform the search. By conveying not only the location and type but also the orientation of objects, it becomes possible to analyze and consider a range of spatial relations that depend on object orientation. To perform OBB detection, we rely on YOLO11, \cite{khanam2024yolov11}, classified as an anchor-free detector, and trained on the DOTA dataset \cite{xia2018dota}. Anchor-free detectors, unlike traditional anchor-based detectors that use a set of fixed-size boxes (anchors) as references across an image, identify objects by predicting key properties like center points or corner keypoints directly from image features processed by a neural network.

\subsection{Relational, Logical Reasoning by Neurosymbolic Inference}

The core of the methodology resides in the reasoning neurosymbolic component, which integrates FOL with probabilistic reasoning. This approach allows us to assign confidence scores to the relationship between objects in the image, based on both their semantic meaning and their spatial configuration, as represented by Oriented Bounding Boxes (OBBs).

\subsubsection{Probabilistic Semiring Reasoning}
\label{subsubsec:prob_sem_reasoning}

The reasoning process is formulated in terms of semiring probabilistic logic \cite{belle2020semiring}, a framework that enables reasoning with soft truth values, where facts, derived from the predicates (such as object labels or spatial relations), are associated with soft truth values in the interval [0,1]. These truth values represent the degree of confidence or probability that a certain fact holds true.

Logical ``AND'' (conjunction) is modeled as the multiplication of probabilities, reflecting the combination of multiple independent pieces of information, where the overall confidence is the product of individual confidence. Similarly, logical OR (disjunction) is modeled as the maximum operation. This is used when multiple possible relations may apply \cite{derkinderen2024semirings}. We select the maximum confidence across all alternatives, reflecting the idea that the truth value of a disjunction is governed by the most probable scenario.

The rules in the system are expressed in Datalog-style logic programs \cite{li2023scallop,bonner1998sequences}, defining the relationships and constraints between objects in the scene. These rules are applied over all combinations of object pairs (detections) in the image, and they help establish how different objects are related to one another according to the query. Table \ref{tab:object_predicates} and Table \ref{tab:predicates} respectively summarize the set of unary object predicates and binary spatial predicates used in our system, together with their formal definitions and semantics. \manu{Table~\ref{tab:predicates} lists the spatial predicates used to express relations between objects. These predicates either originate from the Region Connection Calculus (RCC), using the fuzzy extension proposed by Schockaert et al.~\cite{schockaert2009spatial}, or are user-defined. The application of RCC-based predicates is demonstrated in the flooding scenario presented in Section~\ref{subsec:flooding_scenario}.} 

\begin{table}
\caption{Unary object-type predicates with their corresponding semantics and confidence evaluation. \manu{The objects listed are the entities that can be extracted from the language queries and recognized by the object detector in the images.}}
\label{tab:object_predicates}
\centering
\small  
\begin{tabular}{l|l}
\toprule
\textbf{Predicate} & \textbf{Description} \\ \midrule
\texttt{vehicle} & Confidence that the object is a vehicle. \\ \midrule
\texttt{truck} & Confidence that the object is a truck. \\ \midrule
\texttt{car} & Confidence that the object is a car. \\ \midrule
\texttt{bridge} & Confidence that the object is a bridge. \\ \midrule
\texttt{ship} & Confidence that the object is a ship. \\ \midrule
\texttt{roundabout} & Confidence that the object is a roundabout. \\ \midrule
\texttt{plane} & Confidence that the object is an airplane. \\ \midrule
\texttt{storage\_tank} & Confidence that the object is a storage tank. \\ \midrule
\texttt{baseball\_diamond} & Confidence that the object is a baseball diamond. \\ \midrule
\texttt{basketball\_court} & Confidence that the object is a basketball court. \\ \midrule
\texttt{ground\_track\_field} & Confidence that the object is a ground track field. \\ \midrule
\texttt{harbor}\footnotemark & Confidence that the object is a harbor. \\ \midrule
\texttt{soccer\_ball\_field} & Confidence that the object is a soccer field. \\ \midrule
\texttt{swimming\_pool} & Confidence that the object is a swimming pool. \\ \midrule
\texttt{tennis\_court} & Confidence that the object is a tennis court. \\
\bottomrule
\end{tabular}
\end{table}

\manu{\footnotetext{The entity \emph{harbor} is defined as a small dock for boats, indicated by an oriented bounding box that covers the dock, alongside the land, with the area of water where the boats are located.}}

\begin{table}[h]
{\color{black}
\captionsetup{labelfont={color=black}, textfont={color=black}}
\caption{\manu{The predicates used by the neurosymbolic reasoner can be divided into two categories: (i) \textbf{probabilistic predicates}, which return a soft confidence score in $[0,1]$, and (ii) \textbf{GSD-based predicates}, which return binary values (0 or 1) based on Ground Sample Distance. Among the probabilistic predicates, we distinguish those belonging to the standard RCC framework (shown in \textbf{bold}) from user-defined predicates. User-defined predicates that conceptually overlap with RCC relations are listed in parentheses.}}
\label{tab:predicates}
\centering
\resizebox{\textwidth}{!}{%
\begin{tabular}{l|c|p{12cm}}
\toprule
\textbf{Predicate} & \textbf{Arity} & \textbf{Description} \\
\midrule
\multicolumn{3}{l}{\textbf{Probabilistic predicates}} \\
\midrule
\texttt{(\textbf{DC}, left\_of, right\_of, is\_above, is\_below)} & 2 & A is disconnected from B. According to the implementation of the relations left\_of, right\_of, is\_above, and is\_below, cases in which object A is to the left, right, above, or below object B are included in the DC relation. \\
\midrule
\texttt{\textbf{EC}} & 2 & A is externally connected to B. \\
\midrule
\texttt{\textbf{PO}} & 2 & A is partially overlapping B. \\
\midrule
\texttt{\textbf{TPP}} & 2 & A is a tangential proper part of B. \\
\midrule
\texttt{(\textbf{NTTP}, is\_on)} & 2 & A is a non-tangential proper part of Y. Relation is\_on is equivalent to relation NTTP. \\
\midrule
\texttt{\textbf{EQ}} & 2 & A is equal to B. \\
\midrule
\texttt{\textbf{TTPI}} & 2 & A is a tangential proper part inverse of B. \\
\midrule
\texttt{\textbf{NTTPI}} & 2 & A is a non-tangential proper part inverse of B. \\
\midrule
\texttt{facing\_same} & 2 & A and B face the same direction, based on cosine similarity of their orientation vectors. \\
\midrule
\texttt{facing\_opposite} & 2 & A and B face in opposite directions. \\
\midrule
\texttt{is\_close} & 2 & A and B are spatially close, based on normalized inverse distance. \\
\midrule
\texttt{is\_different} & 2 & A and B are not the same instance. Useful for avoiding self-relations. \\
\midrule
\multicolumn{3}{l}{\textbf{Predicates based on Ground Sample Distance (GSD)}} \\
\midrule
\texttt{is\_close\_meters} & 2 & Returns whether A and B are close based on a physical threshold in meters. \\
\midrule
\texttt{is\_square\_meters} & 1 & Calculates the area of object A in square meters. \\
\bottomrule
\end{tabular}}
}
\end{table}

\subsubsection{Process Flow of the Reasoning Component}
\label{subsubsec:process_flow_reasoning_comp}

The neurosymbolic reasoner operates by taking as input a FOL expression representing the query, alongside the OBBs of the image, which encapsulate the object detections in the scene. The overall workflow proceeds as follows:

The FOL query is first parsed and decomposed into a series of predicates. Each predicate corresponds to a relation between objects, such as "is close", "is left of", or to properties of singular objects such as whether they are of a specific type or \manu{occupy a specific area in terms square meters}. These predicates are then populated with the confidence values derived from the object detections (OBBs). These confidence values reflect the outputs from the earlier stages of the detection process (e.g., the object detection model), indicating the likelihood that a specific object (such as a vehicle, ship, or bridge) is present in the image.

The reasoner generates all possible object tuples, representing combinations of detected objects within the image. Each tuple consists of a pair of objects (detections) that the reasoner will evaluate for the specific relations defined in the query. For each tuple, the corresponding predicate values are retrieved. For example, the predicate "is close" would be evaluated based on the spatial configuration of the two objects, considering their positions and geometries. Geometric features such as distances, angles, and overlaps are used to assess whether the relation described by the predicate holds true. This process involves computing spatial distances, intersection over union (IoU) between bounding boxes, and angular relationships between the objects.

Once all predicates and their corresponding relations are evaluated across the object tuples, the reasoner aggregates the results using probabilistic conjunction. This means applying the ``AND'' operation (multiplication of probabilities) for each predicate, and the ``OR'' operation (max operation) across different possible relations. The final confidence score for the entire query is computed by combining the scores for all predicates, providing a measure of how well the query matches the objects and relations in the image.

This probability reflects the degree to which the query is representative of the image and the most likely configuration of objects and their relationships. This probability serves as an indicator of how well the visual scene, as captured by the object detections and their geometric relations, aligns with the query semantics. A higher probability indicates a stronger match between the query and the image, suggesting that the image corresponds well to the query in terms of both object identity and spatial relationships. To clarify this process, we now present a detailed example of how the system calculates the confidence for a query involving the "left\_of" relation between two objects.

\paragraph{Example} 

Calculation of Confidence for the query \textbf{\emph{a plane to the left of another plane.}} Let's assume we have four objects detected in the image, A, B, C, and D, with the following properties:

\begin{itemize}
    \item Object A: Label = "plane", Confidence = 0.90
    
    \item Object B: Label = "plane", Confidence = 0.85
    
    \item Object C: Label = "plane", Confidence = 0.70
    
    \item Object D: Label = "plane", Confidence = 0.70
\end{itemize}

The query aims at retrieving all images in which there is a plane to the left of another plane. \textbf{Predicate Definition}: The "left\_of" predicate checks whether the x-coordinate of Object A is less than that of Object B. \textbf{Object coordinates}:

\begin{itemize}
    \item Object A has its center at x = 50
    \item Object B has its center at x = 90
    \item Object C has its center at x = 40
    \item Object D has its center at x = 80
\end{itemize}

The confidence score is computed as the product of the object detection confidences, multiplied by either 1 or 0 depending on whether one object is to the left of the other. The left\_of relationship is determined based on the objects’ coordinates. Thus, since the system evaluates the "left\_of" predicate for all the combinations of object pairs in the scene we have:

\begin{itemize}
    \item (A, B): Confidence = $0.90 \cdot 0.85 \cdot 1.00 = 0.77$

    \item (A, C): Confidence = $0.90 \cdot 0.70 \cdot 0.00 = 0.00$ (relation not true)

    \item (A, D): Confidence = $0.90 \cdot 0.70 \cdot 1.00 = 0.63$

    \item (B, C): Confidence = $0.85 \cdot 0.70 \cdot 0.00 = 0.00$ (relation not true)

    \item (B, D): Confidence = $0.85 \cdot 0.70 \cdot 0.00 = 0.00$ (relation not true)

    \item (C, D): Confidence = $0.70 \cdot 0.70 \cdot 1.00 = 0.49$
\end{itemize}

The max operator is then applied to select the highest probability among all pairs. Thus the final probability for the left\_of predicate is: max(0.76, 0.00, 0.63, 0.00, 0.00, 0.49) = 0.76. In more complex queries with additional predicates, the probability is derived by applying the max operator for each predicate and then combining them using multiplication (``AND''). For this example, the "left\_of" relation holds with a confidence of 0.76, which would be used in the reasoning process.

\subsection{Scalable Inference by Factorizing Logic and Conditional Computation}
Complex textual queries can reference up to ten distinct entities. At the same time, in extremely densely populated scenes such as parking lots, object detection may yield hundreds of candidates. Moreover, to minimize FN, a low detection threshold is applied. Consequently, object detection does not significantly reduce the number of candidates given as input to the neurosymbolic reasoning component. Thus, object detection does not significantly contribute to reduce the number of object candidates given in input to the neurosymbolic reasoning component. This leads to an intractable amount of hypotheses.  

Given $\mathbf{N}$ object detections and a query involving $\mathbf{M}$ entities, the total number of possible hypotheses corresponding to assignments of entities to detections is given by $\mathbf{N^M}$. For instance, with $\mathbf{N = 100}$ and $\mathbf{M = 10}$, the number of hypotheses becomes
\[\mathbf{N^M = 100^{10} = (10^2)^{10} = 10^{20}}\] rendering exhaustive inference computationally intractable due to the exponential growth of the hypothesis space.

\begin{figure}[b]
    \centering
    \includegraphics[width=\linewidth]{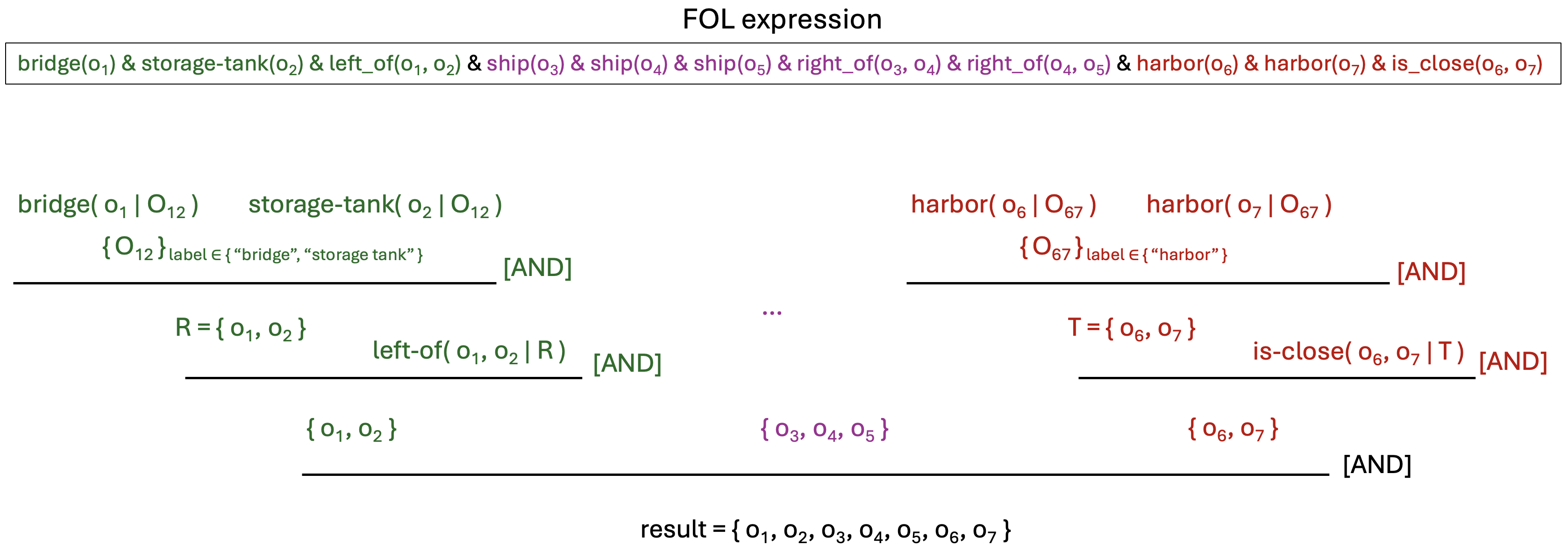}
    \caption{Factorizing the logic into independent clauses (with different colors) reduces the hypotheses significantly. Each clause is computed efficiently on subsets of object detections. Inference becomes computationally tractable.}
    \label{fig:decomposition}
\end{figure}

To address this, we apply a \textbf{clause-level factorization} of the FOL expression derived from the query, factorizing the logic in clauses that can be computed independently of each other, and then combined at the end of the computation. The factorization is illustrated in Figure \ref{fig:decomposition} by decomposition clauses, that are grouped and color-coded according to their independence. For instance, a clause expressing that a bridge is \emph{left of} a storage tank can be evaluated separately from a clause asserting that two harbors are \emph{close to} each other.

In addition to structural factorization, we apply \textbf{conditional computation} over the subset of detections relevant to each clause. Suppose the image contains $\mathbf{N = 100}$ detected objects, including 10 bridges, 5 storage tanks, 4 harbors, 10 ships, and 71 miscellaneous objects (e.g. bridges). Then the first clause (bridge \emph{left of} storage tank) involves $\mathbf{10 \times 5}$ hypotheses, the second clause (e.g., a ternary relation involving ships) considers $\mathbf{10 \times 10 \times 10}$ hypotheses, and the third clause (harbor \emph{near} harbor) involves $\mathbf{4 \times 4}$ hypotheses. This results in a total number of hypotheses: \[\mathbf{10 \times 5 + 10 \times 10 \times 10 + 4 \times 4 = 50 + 1000 + 16 = 1066}\] which is many orders of magnitude smaller than the $\mathbf{10^{20}}$ combinations. The conditional computation is shown in Figure \ref{fig:decomposition} by the subsets $\{O_i\}$ in which clauses are evaluated. This combination of clause-level factorization and detection-level filtering enables tractable inference over rich and expressive queries.

\section{Experimental Setup}
\label{sec:experiments}

\manu{
\subsection{Technical goals}
\label{subsec:technical_goals_limitations}
The goal of our framework is to ensure that images can be retrieved even when accurate metadata—typically produced during the standard processing steps of remote sensing datasets—is missing, either due to technical glitches or human errors.
Our framework currently supports both Red-Green-Blue (RGB) and grayscale images. The RGB images are sourced from the DOTA dataset \cite{Long2021DiRS}, derived from Google Earth and Cyclomedia, while the grayscale images originate from the GF-2 and JL-1 satellites. The image sizes in the dataset range from approximately 800 × 800 to 4000 × 4000 pixels.
The images used to test our methodology are characterized by a maximum Ground Sample Distance (GSD) of 2.25, a minimum GSD of 1.34e-06, and an average GSD of 0.39. The spatial resolution ranges between 0.50 m and 3.20 m, while the radiometric resolution varies between 10 and 11 bits. Our methodology is designed to analyze orthogonal imagery and has not been tested on oblique images. As the baselines we compare against are foundation models, we empirically observe that most current models perform well when the smallest object dimension (width or height) is at least 40 pixels.
}

\subsection{Dataset}

To test our methodology we repurpose \cite{Long2021DiRS} the DOTA dataset \cite{xia2018dota}, which comprises many satellite images with various semantic contents. We build on DOTA because it has ground truth objects (OBBs), by which we can evaluate explainability in terms of the predicted objects that match the query at hand. Existing text-image retrieval datasets are characterized by text-image pairs without any such OBB ground truths. We transform the DOTA dataset into a text-to-image retrieval dataset by captioning its images with five queries each, from simple to complex semantics. We evaluate how increasing query complexity affects the retrieval performance of each model.

\subsubsection{Query Complexity}

Complexity is defined by these indicators: 

\begin{itemize}
    \item \textbf{Number of objects}: each query is characterized by a specific number of entities described. 
    \item \textbf{Number of types of objects}: the entities in each query belong to specific object categories, and the number of these categories varies depending on the query.
    \item \textbf{Relationship complexity}: the relations between objects defined in the queries can involve a varying number of objects. The higher the number of objects involved in a relation, the higher the complexity of the relation itself. 
\end{itemize}

\begin{figure}[h]
    \centering
    \includegraphics[width=\linewidth]{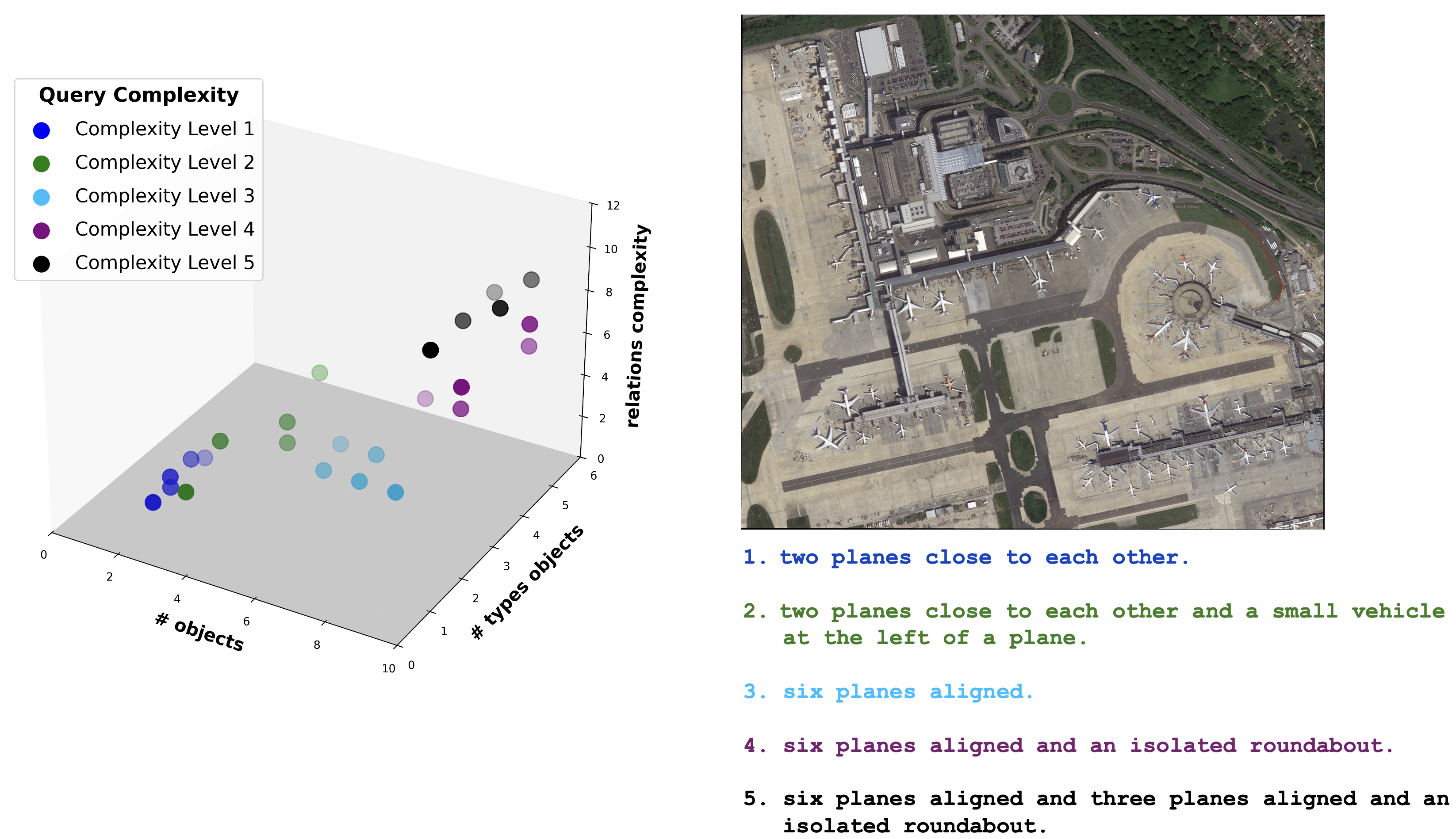}
    \caption{The complexity of queries is defined by the amount of objects and types and the complexity of the relationships between them. The left panel quantifies this for various query complexity levels. For an example image, the right panel shows five queries at increasing levels of complexity.}
    \label{fig:query_complexity}
\end{figure}

For each of the three factors defining the complexity of the query, the higher the number of each of them, the higher the complexity of the query used. Figure \ref{fig:query_complexity} shows the graphical representation of query complexity decomposition. On the left, the three-dimensional graph assigns each axis to a specific complexity dimension: the number of objects in the query, the number of object types, and the relation complexity—defined as the cumulative complexity of the individual relations within the query. Queries are categorized into five levels, from Level 1 (minimal complexity) to Level 5 (maximum complexity). On the right, an example image is displayed along with its corresponding queries, demonstrating how query complexity varies across levels. Color coding is used to indicate the cluster in the three-dimensional graph to which each query belongs.

\subsubsection{More Complexity than Existing Datasets} 

A fundamental reason for which we do not rely on existing text-image retrieval datasets is their lack of complexity. To show the complexity difference, we provide queries from the RSICD \cite{lu2017exploring} and RSITMD \cite{yuan2022exploring} datasets, and compare their complexity to our dataset. 

Queries belonging to the RSICD dataset: 

\begin{itemize}
    \item A playground is between a baseball field and a large building.
    \item Some planes are in an airport surrounded by many green trees.
    \item A football field and five basketball fields are surrounded by some green trees. 
\end{itemize}

Queries belonging to the RSITMD dataset: 

\begin{itemize}
    \item The baseball field is next to a basketball field, a volleyball court and a tennis court.
    \item There are three white planes parked in the grey parking lot.
    \item There are two baseball fields, near the lawn, next to the basketball court and parking lot.
\end{itemize}

Queries belonging to the DOTA dataset, one for each complexity level: 

\begin{enumerate}
    \item A storage tank \textcolor{blue}{to the left of} another storage tank.
    \item A bridge \textcolor{blue}{to the left of} a tank and two ships close to each other.
    \item A bridge \textcolor{blue}{to the left of} a tank and six tanks \textcolor{red}{in column}.
    \item There are six planes \textcolor{red}{aligned} and an isolated roundabout.
    \item Four tennis courts \textcolor{red}{aligned} and three tennis courts \textcolor{red}{aligned} and a soccer ball field inside a ground track field. 
\end{enumerate}

\begin{figure}[h]
    \centering
    \includegraphics[width=0.70\linewidth]{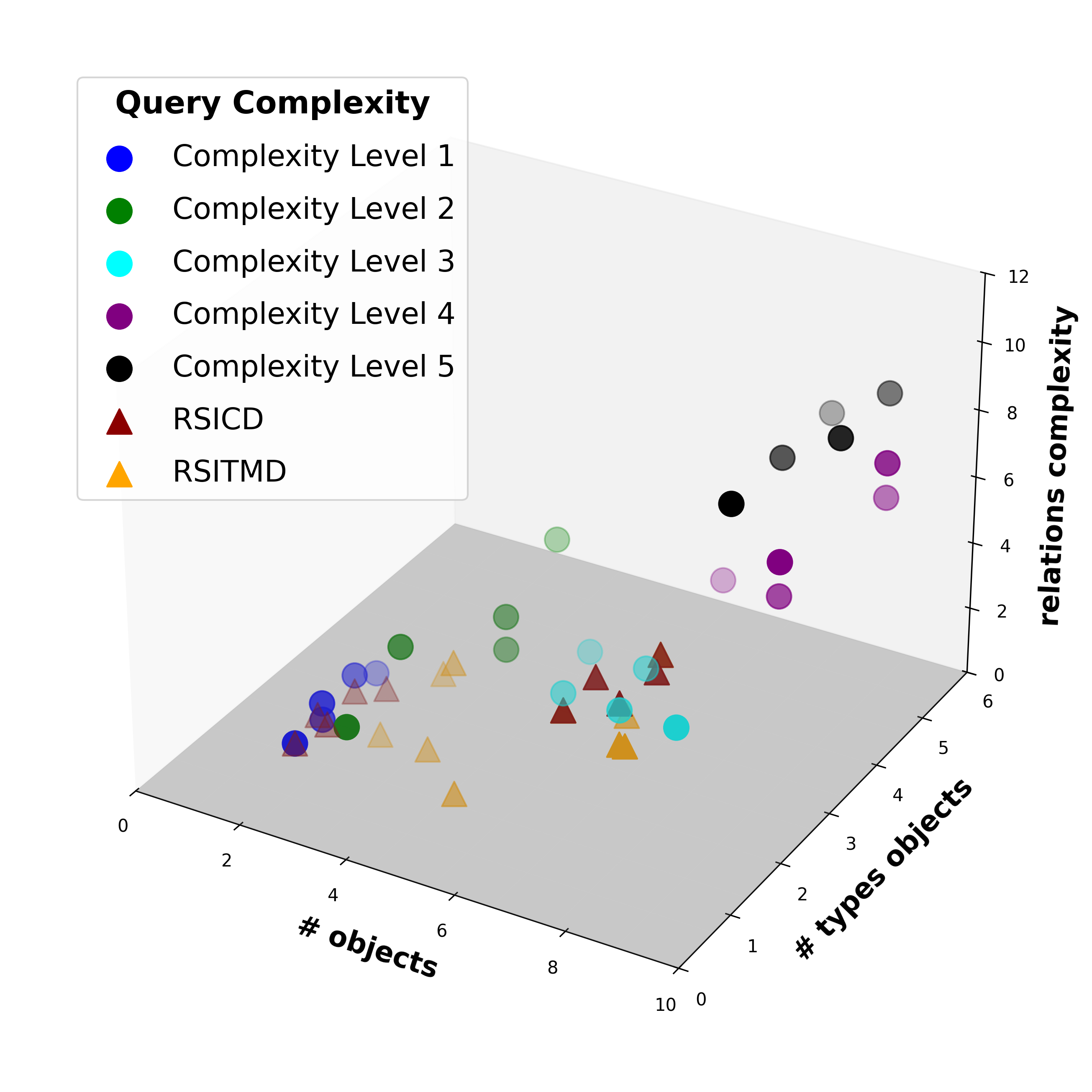}
    \caption{Our queries are more complex than the queries from the RSICD \cite{lu2017exploring} and RSITMD \cite{yuan2022exploring} datasets, especially of the queries of level four and level five.}
    \label{fig:graph_comparison}
\end{figure}

The examples highlight the differences between the queries in the repurposed DOTA dataset and those in the RSICD and RSITMD datasets. RSICD and RSITMD queries match DOTA queries in terms of the number and variety of objects up to complexity level three. However, at complexity levels four and five, our dataset clearly surpasses both RSICD and RSITMD. Our queries often include up to ten objects, whereas the RSICD and RSITMD datasets are generally limited to two–five objects and two–four object classes.

More importantly, the superior complexity of our queries lies in their relational structure. As illustrated in the examples, our queries consistently include relations that convey \textcolor{blue}{orientation}, which go beyond mere proximity to suggest specific spatial behaviors (e.g., to the left of vs. near). Additionally, beginning at complexity level three, our queries feature \textcolor{red}{non-atomic} relations, involving more than two objects. These relations, which are later decomposed into corresponding atomic components by the neurosymbolic system, significantly increase the overall complexity. They not only involve multiple objects simultaneously but also incorporate both orientation and proximity, further enriching the semantic depth of the queries.

The difference in query complexity is clearly illustrated in Figure \ref{fig:graph_comparison}, which breaks down query complexity into three components: number of objects, number of object types, and relational complexity. While RSICD and RSITMD queries are comparable to ours in terms of the number of objects and object classes (up to complexity level three), a significant gap emerges when relational complexity is considered, highlighting the superior expressiveness of our queries in this aspect.

\manu{
\subsubsection{Image Uncertainty} 
\label{subsubsec:image_uncertainty}
In the DOTA dataset, each object in an image is assigned a difficulty level for detection, where 0 indicates an easy-to-detect object and 1 indicates a difficult-to-detect object. Table \ref{tab:difficulty} shows the detection difficulty for each object in the dataset. 

\manu{
\begin{table}[ht]
\centering
{\color{black}
\captionsetup{labelfont={color=black}, textfont={color=black}}
\caption{Table presenting the average detection difficulty of each object category across all images. Storage tank emerges as the most challenging to detect, whereas ship is the easiest.}
\label{tab:difficulty}
\begin{tabular}{l|c}
\toprule
\textbf{Object} & \textbf{Detection Difficulty} \\
\midrule
storage tank & 0.34 \\
soccer ball field & 0.29 \\
swimming pool & 0.15 \\
bridge & 0.08 \\
ground track field & 0.08 \\
roundabout & 0.08 \\
small vehicle & 0.06 \\
basketball court & 0.06 \\
baseball diamond & 0.04 \\
plane & 0.03 \\
tennis court & 0.03 \\
large vehicle & 0.02 \\
harbor & 0.01 \\
ship & 0.01 \\
\bottomrule
\end{tabular}
}
\end{table}
}

Other than calculating the average level of uncertainty for each object, we can define a property named Image Uncertainty (IU),  which for each image defines the uncertainty characterizing and that is derived from its object. We define IU as follows: 

\begin{align}
\label{eq:iu}
    IU = \frac{1}{n}\sum_{i = 0}^{n}difficulty(object_{i})
\end{align}

where $n$ is the number of objects in the image and $difficulty(object_{i}) \in \{0, 1\}$ indicates the detection difficulty of object $i$.

A higher IU corresponds to a greater uncertainty regarding the object detector's performance on that image and thus the final retrieval performance. The value of IU is between zero and one. Figure \ref{fig:image_uncertainty_example} illustrates two examples of Image Uncertainty.

\begin{figure}[h]
    \centering
    \includegraphics[width=0.90\linewidth]{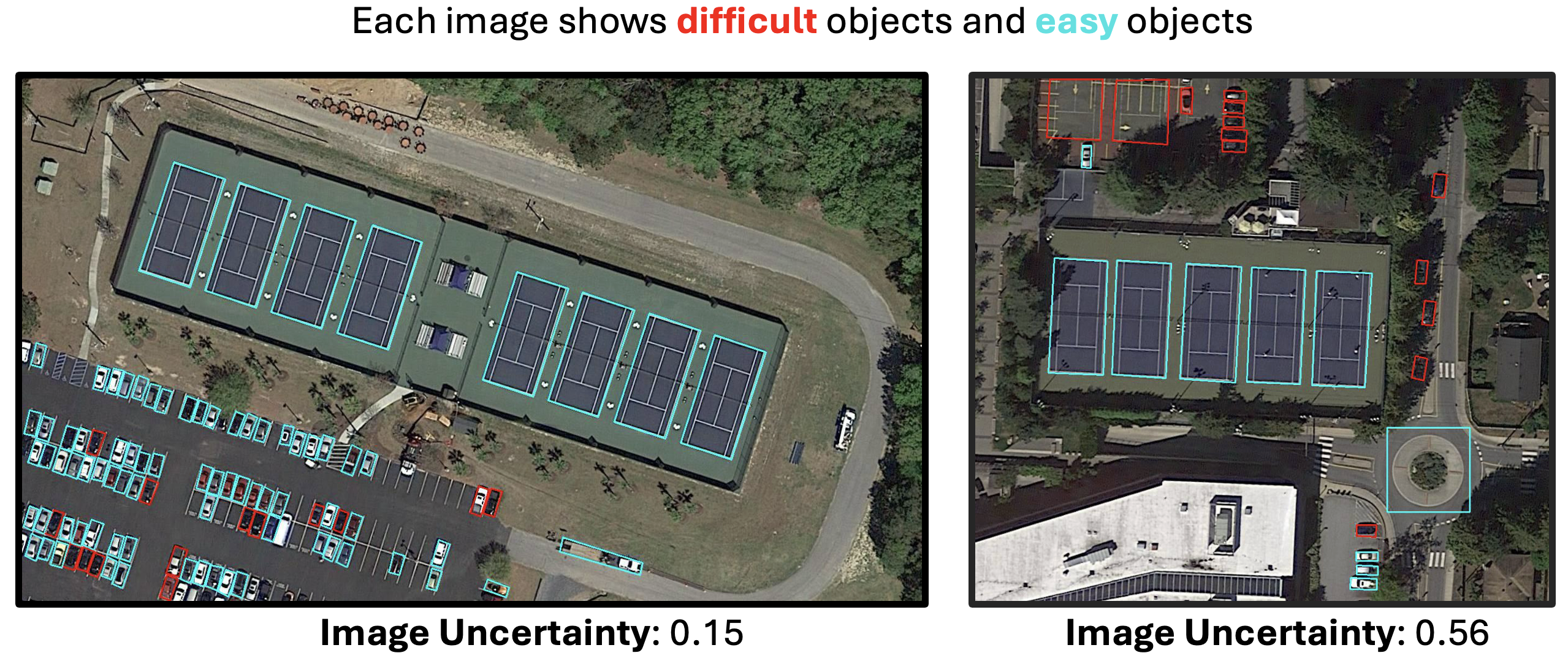}
    \caption{Images characterized by different levels of uncertainty: the image on the left presents more objects which are difficult to detect, while the image on the right presents more object which are easy to detect.}
    \label{fig:image_uncertainty_example}
\end{figure}
}

\subsection{Evaluation Metrics}

To evaluate the model performance, we use the standard evaluation metrics in retrieval tasks and measure the rank-based performance by R@k, P@k, mean recall (mR), and mean precision (mP) \cite{hoxha2020toward}, \cite{yuan2022exploring}. With different values of k, R@k means the fraction of queries for which the most relevant item is ranked among the top-k retrievals, while P@k means the proportion of the top-k retrieved (or recommended) items that are relevant. mR and mP calculate the average performance across all the k values respectively for recall and precision. We report the results of k = [1, 5, 10], as in previous works. 

\subsubsection{Robustness to Semantic Complexity}
Moreover, for both R@k and P@k, we calculate how increasing the complexity of the query impacts P@k and R@k. To accomplish this, we define the metric \emph{Retrieval Robustness to Query Complexity (RRQC)}: 

\begin{align}
RRQC_{m@k, d} = \frac{1}{|S_d|} \sum_{\substack{i, j \in \{1, 2, 3, 4, 5\} \\ i < j \\ j - i = d}} \left( m@k_i - m@k_j \right)
\end{align}

where $m@k$ ($P@k$ or $R@k$) represents the metric of interest, $d \in \{1, 2, 3, 4\}$ is the chosen distance between query complexity levels, $i$ and $j$ represent two different query complexity levels at which we consider the metric of interest, and $|S_d|$ represents the number of $(i, j)$ pairs. 

\manu{
\subsubsection{Robustness to Image Uncertainty}
\label{subsubsection:riu}
In addition to evaluating robustness to query semantic complexity, we analyze how increasing detection uncertainty of objects in images affects retrieval performance. To quantify this effect, we introduce the metric \emph{Retrieval Robustness to Image Uncertainty (RRIU)}.

For each image, we calculate the hit ratio, defined as the proportion of times it appears in the top-k positions of the ranking relative to its occurrences in the ground truth:

\begin{align}
    \hat{p_{i}} = \frac{N^{(k)}_{Hit}(i)}{N_{GT}(i)}
\end{align}

where $N^{(k)}_{Hit}(i)$ is the number of times the image is retrieved within the top-k, and $N_{GT}(i)$ is the number of times it is relevant. To obtain a single representative value per image across multiple cutoffs, we average over the selected $k$ values:  

\begin{align}
    \overline{p}_{i} = \frac{1}{K}\sum_{k \in K}\hat{p_{i}}^{(k)}, \: K = \{1, 5, 10\}
\end{align}

Images are then partitioned into $M$ difficulty bins $B_{1}, ... , B_{M}$, based on their $IU$ as defined in Equation \ref{eq:iu}. The average ratio within each bin is: 

\begin{align}
    \hat{P}(B_{j}) = \frac{1}{|B_{j}|}{\sum_{i \in B_{j}}\overline{p}_{i}}
\end{align}

Finally, the RRIU between two bins is defined as the difference in their average hit ratios, reflecting the impact of image uncertainty on retrieval performance: 

\begin{align}
    RRIU_{B_{i}, B_{j}} = P(B_{j}) - P(B_{i}), \text{where} \: i, j \in \{1 ,...,  M\} \: \text{and} \: i < j
\end{align}
}

\subsection{Implementation}
To perform the text-to-logic conversion we employ GPT-4o \cite{hurst2024gpt} which is the OpenAI state-of-the art model at the time of writing. We perform oriented bounding box object detection, by employing YOLO11 \cite{khanam2024yolov11}. We implement the neurosymbolic component by relying on Scallop \cite{li2023scallop}. We run the LVLMs relying on one NVIDIA-L4 GPU. 

\definecolor{lightred}{RGB}{255,100,100}

\manu{
\subsection{LLMs proficiency in text-to-logic translation.}
\label{subsection:llms_profiency_logic} 
The proficiency of LLMs in text-to-logic translation has been shown by Yang et al. \cite{yang2023harnessing}. They evaluated the ability to translate natural language queries into FOL expressions using the FOL Bilingual Bilingual Evaluation Understudy (BLEU) and FOL Logical Equivalence (LE) metrics. Here, we quantify LLM proficiency using the same metrics in our scenarios and across different query complexity levels. Results are shown in Table \ref{tab:text_logic_translation}. For both FOL BLEU and FOL LE, performance is perfect for the first two levels of query complexity. For levels three, four, and five, performance decreases, ranging from 100\% to 94\% for FOL BLEU and to 99\% for FOL LE. These results are consistent with those reported in related literature and confirm that LLMs can perform reliable text-to-logic translation in the context of RS.
}

\begin{table}[h]
\centering
{\color{black}
\captionsetup{labelfont={color=black}, textfont={color=black}}
\caption{GPT-4o achieves 100\% on FOL BLEU and FOL LE for queries of complexity levels 1 and 2, with performance slightly decreasing to 0.94 (FOL BLEU) and 0.99 (FOL LE) for complexity levels three to five.}
\label{tab:text_logic_translation}
\begin{tabular}{llllll}
\hline
                  & \multicolumn{5}{c}{\textbf{Query Complexity Level}}                                          \\ \hline
                  & \textbf{Level 1} & \textbf{Level 2} & \textbf{Level 3} & \textbf{Level 4} & \textbf{Level 5} \\ \hline
\textbf{FOL BLEU} & 1.00             & 1.00             & 0.94             & 0.94             & 0.94             \\
\textbf{FOL LE}   & 1.00             & 1.00             & 0.99             & 0.99             & 0.99             \\ \hline
\end{tabular}
}
\end{table}

\section{Comparison with RS-LVLMs}
\label{section:comparison_with_rs_lvms}

We compare the proposed method RUNE with three general-purpose LVLMs and three RS-LVLMs. 

\begin{itemize}
    \item \textbf{CLIP (Contrastive Language–Image Pretraining)} \cite{radford2021learning}: is a multimodal model by OpenAI that learns joint image and text representations using contrastive learning on image–text pairs, enabling zero-shot transfer to vision-language tasks.
    \item \textbf{OpenCLIP} \cite{cherti2023reproducible}: an open-source implementation of CLIP, extending its architecture and training on large-scale datasets such as LAION-400M to support broad multimodal research and applications.
    \item \textbf{LlaVA (Large Language and Vision Assistant)} \cite{liu2023visualinstructiontuning}: a multimodal model that integrates a vision encoder with a large language model to enable visual instruction following. It is trained on image–text pairs and vision-language instruction data to support tasks such as visual question answering and image-grounded dialogue.
\end{itemize}

Remote Sensing Large Visual Language Models: 

\begin{itemize}
    \item \textbf{RemoteCLIP} \cite{liu2024remote}: a vision-language foundation model specifically designed for RS imagery. It is trained on a large-scale dataset of RS images and aligned textual descriptions, enabling it to learn robust visual features with rich semantic information and aligned text embeddings.
    \item \textbf{RS-LLaVA} \cite{bazi2024rs}: a large vision-language model tailored for RS. RS-LLaVA is trained in multiple stages to align visual and textual features and to follow instructions related to RS tasks.
    \item \textbf{GeoRSCLIP} \cite{zhang2024rs5m}: a vision-language model for RS that builds upon the CLIP (Contrastive Language-Image Pre-training) architecture. It is fine-tuned or adapted using RS image-text paired data to enhance its understanding of geospatial visual concepts and their corresponding textual descriptions. 
\end{itemize}

\begin{table}[b]
\caption{Retrieval performance on the repurposed DOTA dataset for the first level of complexity query. RUNE structurally outperforms the state-of-the-art RS-LVLMs in terms of recall (R) as well as precision (P).}
\label{tab:retrieval_results}
\resizebox{\textwidth}{!}{%
\begin{tabular}{lll|lll|l|lll|l}
\toprule
\textbf{Model} & \textbf{Image Backbone} & \textbf{Text Backbone} & \textbf{R@1} & \textbf{R@5} & \textbf{R@10} & \textbf{mR} & \textbf{P@1} & \textbf{P@5} & \textbf{P@10} & \textbf{mP} \\
\midrule
RUNE (ours) & YOLO11 & GPT-4o & \cellcolor[HTML]{9AFF99}0.12 & \cellcolor[HTML]{9AFF99}0.35 & \cellcolor[HTML]{9AFF99}0.53 &
\cellcolor[HTML]{9AFF99}0.33 & \cellcolor[HTML]{9AFF99}1.00 & \cellcolor[HTML]{9AFF99}0.87 & \cellcolor[HTML]{9AFF99}0.78 & \cellcolor[HTML]{9AFF99}0.88 \\ \midrule
\multicolumn{11}{c}{\textbf{RS-LVLMs}} \\ \midrule
\multirow{-1}{*}{RemoteCLIP \cite{liu2024remote}} 
 & ViT-B-32 & Transformer & 0.07 & 0.27 & 0.44 & 0.26 & 0.63 & 0.69 & 0.61 & 0.64 \\
 & ViT-L-14 & Transformer & 0.07 & 0.29 & 0.42 & 0.26 & 0.63 & 0.64 & 0.55 & 0.61 \\
 & RN50      & Transformer & 0.06 & 0.24 & 0.42 & 0.24 & 0.50 & 0.57 & 0.57 & 0.55 \\
GeoRSCLIP \cite{zhang2024rs5m} & ViT-B-32 & Transformer & 0.06 & 0.20 & 0.32 & 0.19 & 0.50 & 0.49 & 0.46 & 0.48 \\
RS-LLaVA \cite{liu2024remote} & CLIP-ViT-L-336px & Mistral-7B & 0.03 & 0.17 & 0.24 & 0.15 & 0.53 & 0.48 & 0.35 & 0.45  \\ \midrule
\multicolumn{11}{c}{\textbf{General Purpose LVLMs}} \\ \midrule
\multirow{-1}{*}{OpenCLIP \cite{liu2024remote}} 
 & ViT-L-14 & Transformer & 0.08 & 0.22 & 0.36 & 0.22 & 0.80 & 0.61 & 0.56 & 0.66 \\
 & ViT-B-32 & Transformer & 0.04 & 0.21 & 0.33 & 0.20 & 0.70 & 0.57 & 0.48 & 0.58 \\
 & RN50     & Transformer & 0.06 & 0.18 & 0.29 & 0.18 & 0.70 & 0.54 & 0.47 & 0.57 \\ 
CLIP \cite{radford2021learning} & ViT-B-32 & Transformer & 0.04 & 0.21 & 0.33 & 0.20 & 0.70 & 0.57 & 0.48 & 0.58 \\
LLaVA-1.6 & CLIP-ViT-L-336px & GPT-4 & 0.02 & 0.11 & 0.17 & 0.10 & 0.43 & 0.37 & 0.30 & 0.37 \\
\bottomrule
\end{tabular}
}
\end{table}

\begin{figure}[t]
  \centering
  \includegraphics[width=\linewidth]{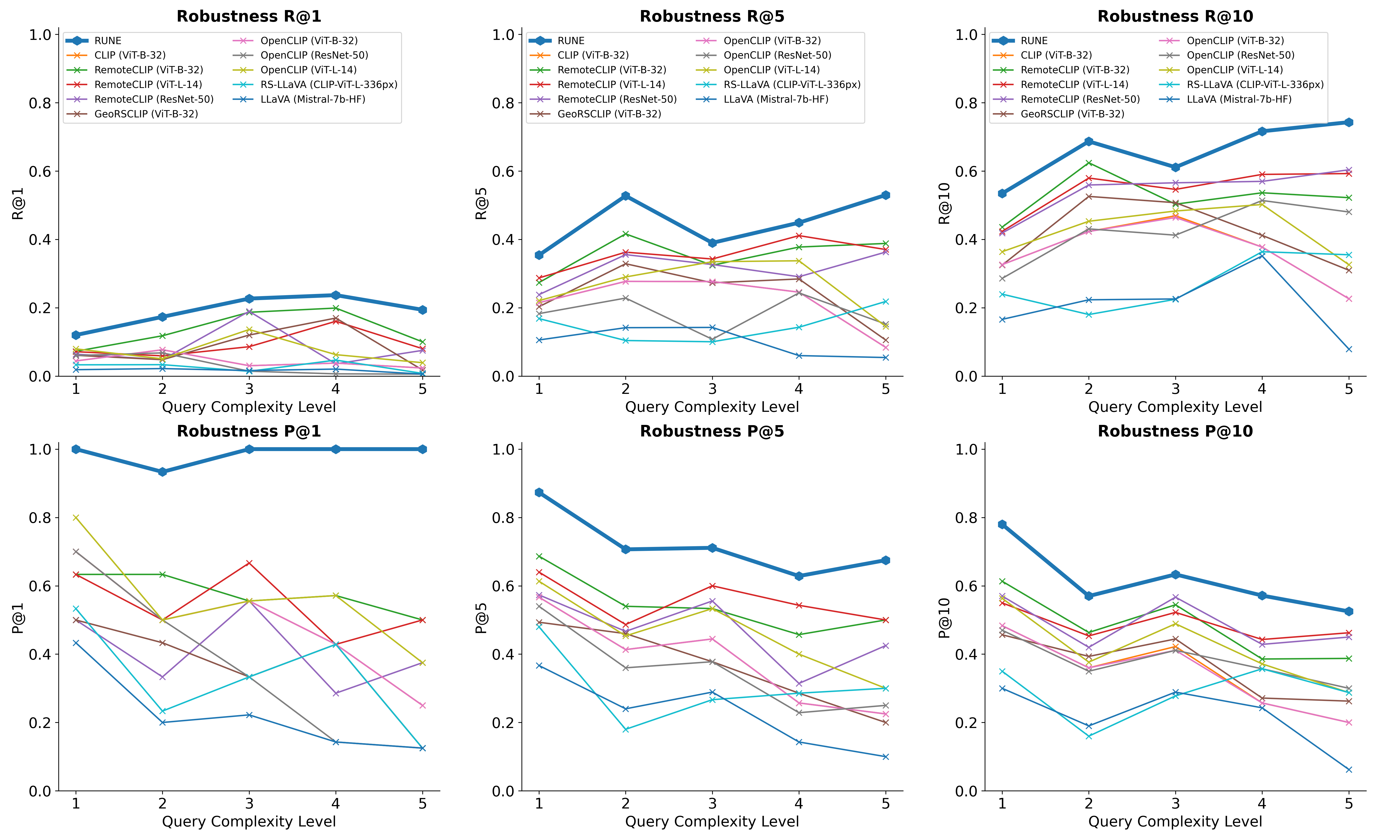}
  \caption{The text-image retrieval performance of RUNE is structurally better than the state-of-the-art RS-LVLM models, especially for more complex queries, for precision as well as recall.}
  \label{fig:level_change}
\end{figure}

\subsection{R@k and P@k}
Table \ref{tab:retrieval_results} shows the results related to text-image retrieval for queries of complexity level one. Our method surpasses the LVLMs in both P@k and R@k. Noteworthy is the P@1 in which our method registers 100\% thus denoting the absence of False Positives. Focusing on the models specialized in RS, RUNE, with an mR of 0.33 and mP of 0.88, outperforms RemoteCLIP, by 23.73\% in mR and 31.58\% mP. RS-LLaVA on the contrary is the model that performs the worst, with a mR of 0.15 and a mP of 0.45. 

Focusing on the generalist models OpenCLIP (ViT-L-14) and CLIP (ViT-B-32) are the models with the highest performance with a mR of 0.20 and a mP of 0.66. LLaVA-1.6 instead is the generalist model and the model in general with the worst performance with an mR of 0.10 and an mP of 0.37. 

Overall, RUNE overcomes in terms of P@K and R@K all the models both the ones fine-tuned on RS data and the generalist models. 

\subsection{Robustness}
In figure \ref{fig:level_change} we analyze how the performance varies when varying the complexity level of the query used to retrieve the target images. For R@5 and R@10 RUNE is characterized by improved performance when the complexity of the query increases, thus indicating enhanced robustness. All the other models apart from RemoteCLIP in the case of R@5 and RemoteCLIP and GeoRSCLIP for R@10 are characterized by a sharp decline in performance. Noteworthy is the trend observed in the P@K. RUNE shows almost perfect stability in its P@1 which is always equal to 100\% apart for a query complexity level of two. For P@5 and P@10, even though growing the query complexity level brings a decrease in the performance RUNE for each complexity level RUNE beats all the other models, with superior P@5 and P@10. 

\begin{table}[t]
\centering
\caption{Robustness of RUNE and the state-of-the-art RS-LVLMs, where a positive number means an increased performance for increased query complexity, and a negative means a performance decrease. Metric RRQC is explained in the text. RUNE is the most robust model, while all the other models show a significant performance degradation for increasing query complexity.}
\label{tab:performance_change}
\resizebox{0.70\textwidth}{!}{%
\begin{tabular}{lllll}
\toprule
\textbf{Model} & \textbf{Image Encoder} & \textbf{Text-encoder} & $\mathbf{RRQC_{R@k,4}}$ & $\mathbf{RRQC_{P@k,4}}$ \\
\midrule
RUNE (ours) & YOLO11 & GPT-4o & \cellcolor[HTML]{9AFF99}9.96\% & \cellcolor[HTML]{9AFF99}-5.39\% \\ \midrule
\multicolumn{5}{c}{\textbf{RS-LVLMs}} \\ \midrule
\multirow{-1}{*}{RemoteCLIP \cite{liu2024remote}} 
 & ViT-L-14 & Transformer & 5.86\% & -5.45\% \\
 & ViT-B-32 & Transformer & 7.06\% & -8.34\% \\
 & RN50      & Transformer & 7.80\% & -6.82\% \\
RS-LLaVA \cite{liu2024remote} & CLIP-ViT-L-336px & Mistral-7B & -4.88\% & -15.82\%  \\
GeoRSCLIP \cite{zhang2024rs5m} & ViT-B-32 & Transformer & -14.74\% & -21.55\% \\ \midrule
\multicolumn{5}{c}{\textbf{General Purpose LVLMs}} \\ \midrule
\multirow{-1}{*}{OpenCLIP \cite{liu2024remote}} & ViT-B-32 & Transformer & -15.39\% & -22.00\% \\ 
CLIP \cite{radford2021learning}  
 & ViT-L-14 & Transformer & -10.02\% & -17.15\% \\
 & ViT-B-32 & Transformer & -15.39\% & -22.00\% \\
 & RN50      & Transformer & -11.08\% & -21.41\% \\ 
LLaVA-1.6 & CLIP-ViT-L-336px & Mistral-7b-HF & -20.13\% & -29.65\% \\
\bottomrule
\end{tabular}
}
\end{table}

\subsection{Complex Queries}

\begin{figure}[h]
    \centering
    \includegraphics[width=\linewidth]{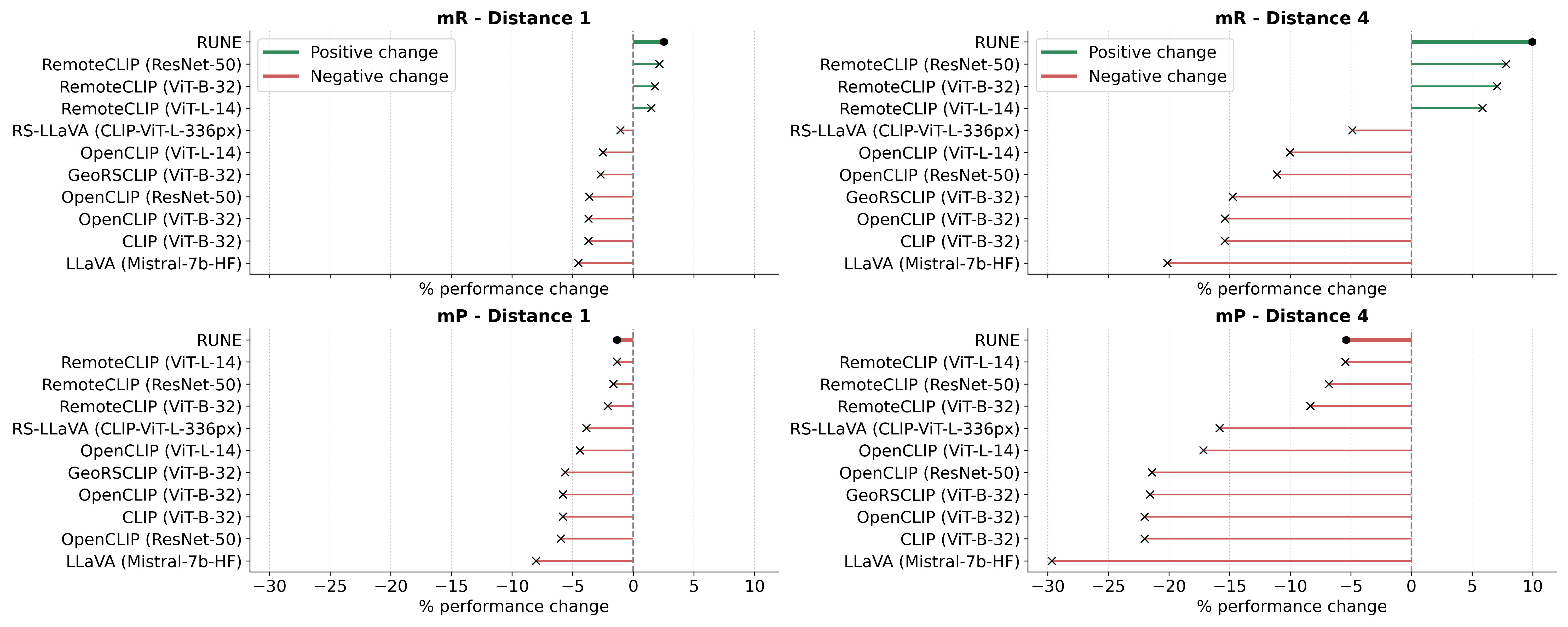}
    \caption{RUNE is more performant and robust than the state-of-the-art RS-LVLM models, for mR as well as mP. The change from simple to more complex queries is indicated by the distance, i.e. the change of complexity levels. For an increase in complexity, the performance of RUNE is most stable, i.e. it degrades the least.}
    \label{fig:performance_change}
\end{figure}

Table \ref{tab:performance_change} and Figure \ref{fig:performance_change} show how the mR of the different models change when raising the complexity of the queries. Table \ref{tab:performance_change} shows that the only robust model that keeps its performance constant or also increases it is RUNE, while all the other models decrease in performance. The highest performance decrease for mR is seen with the OpenCLIP model which is followed by the LLaVA-1.6 model. RemoteCLIP, despite a reduction in performance with increased query complexity, experiences a decrease that is a third of the one observed with OpenCLIP and a fourth of the decrease of LLaVA-1.6. The backbone with the highest robustness for RemoteCLIP is ViT-L-14. Figure \ref{fig:performance_change} illustrates how both mR and mP vary as a function of query complexity, analyzed across complexity distances one and four. We define complexity distance as the absolute difference between two levels of query complexity used to compute pairwise performance changes. A complexity distance of 1 indicates comparisons between immediately adjacent levels (e.g., level 1 vs. level 2, level 2 vs. level 3, etc.), while distances of 2 or more capture differences across increasingly dissimilar levels (e.g., level 1 vs. level 3 for distance 2, level 1 vs. level 4 for distance 3, and so on). This formulation allows us to quantify how performance trends evolve not only with increasing complexity as well as with the relative disparity in complexity between queries. 

Focusing on the mR, we observe that RUNE and RemoteCLIP are the only models which show a positive change when varying the complexity level. This counterintuitive improvement can be attributed to a key factor: more complex queries include additional relational constraints and object-specific details, which help reduce semantic ambiguity. For a model like RUNE, which is designed to leverage compositional and relational semantics, these constraints act as strong semantic anchors. They guide the retrieval system toward a more precise interpretation of the query, increasing the likelihood of retrieving a more relevant set of items—thereby improving recall.

All the other models instead show a negative change, with the worst model represented by LLaVA with a negative change between query complexity level 1 and query complexity level 5 of -20.13\%. The graphs related to the percentage change in the precision present a similar trend. While for distance level 1 RemoteCLIP (ViT-L-14) and RUNE show the same negative change (-1.35\%), when query distance is 4, thus the change between query complexity level 1 and query complexity level 5 RUNE show inferior negative change with -5.39\% compared to -5.45\% of RemoteCLIP (ViT-L-14). 

\manu{
\subsection{Robustness to Image Uncertainty}
\label{subsec:image_uncertainty_results}
Figure \ref{fig:uncertainty_comparison} illustrates how model performance varies with increasing levels of uncertainty in remote sensing images, transitioning from an "easy" detection level to a "hard" detection level.
As expected, all models exhibit a decline in performance under increased uncertainty. The model with the smallest performance drop is RemoteCLIP (ViT-L-14), with a decrease of 8.30\%, followed closely by GeoRS-CLIP, which shows a decrease of 8.36\%. In contrast, the model most affected by uncertainty is LLaVA (Mistral-7b-HF), which suffers a substantial performance decline of 44\%.

Our proposed model, RUNE, falls into the second-best performance bracket, with a decrease of 15.17\%. This is a smaller drop compared to RemoteCLIP (ResNet-50), which shows a decline of 16.70\%.

The performance decrease of RUNE can be attributed to its reliance on object detection for retrieval. Unlike large vision-language models (LVLMs), which directly encode the entire image into an embedding space and may therefore generalize better under uncertain or degraded image conditions, RUNE’s performance is more sensitive to detection difficulty. Nonetheless, RUNE still outperforms six out of the LVLMs in terms of robustness to image uncertainty, indicating that its detection-based, probabilistic reasoning approach holds strong potential under realistic remote sensing scenarios.

\begin{figure}
    \centering
    \includegraphics[width=\linewidth]{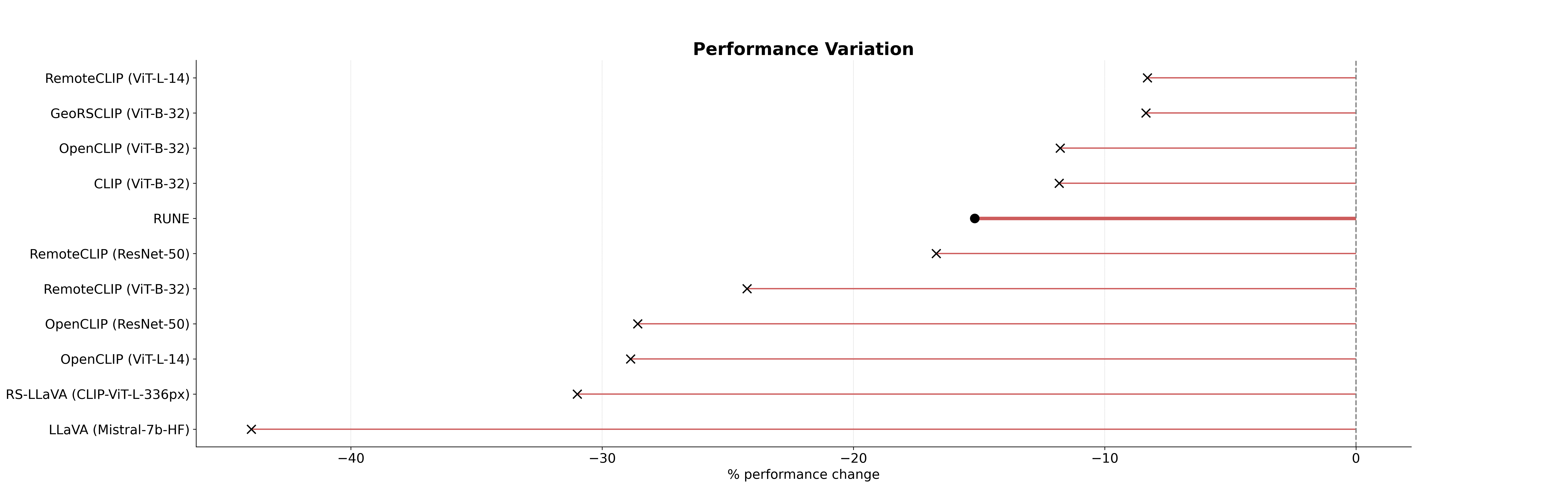}
    \caption{Every model suffers the impact of increased object uncertainty on its performance. RUNE position itself in the second bracket, with a performance decrease of 15\%.}
    \label{fig:uncertainty_comparison}
\end{figure}
}

\subsection{Explainability}
\begin{figure}
     \centering
     \begin{subfigure}{\textwidth}
         \centering
         \includegraphics[width=0.90\textwidth]{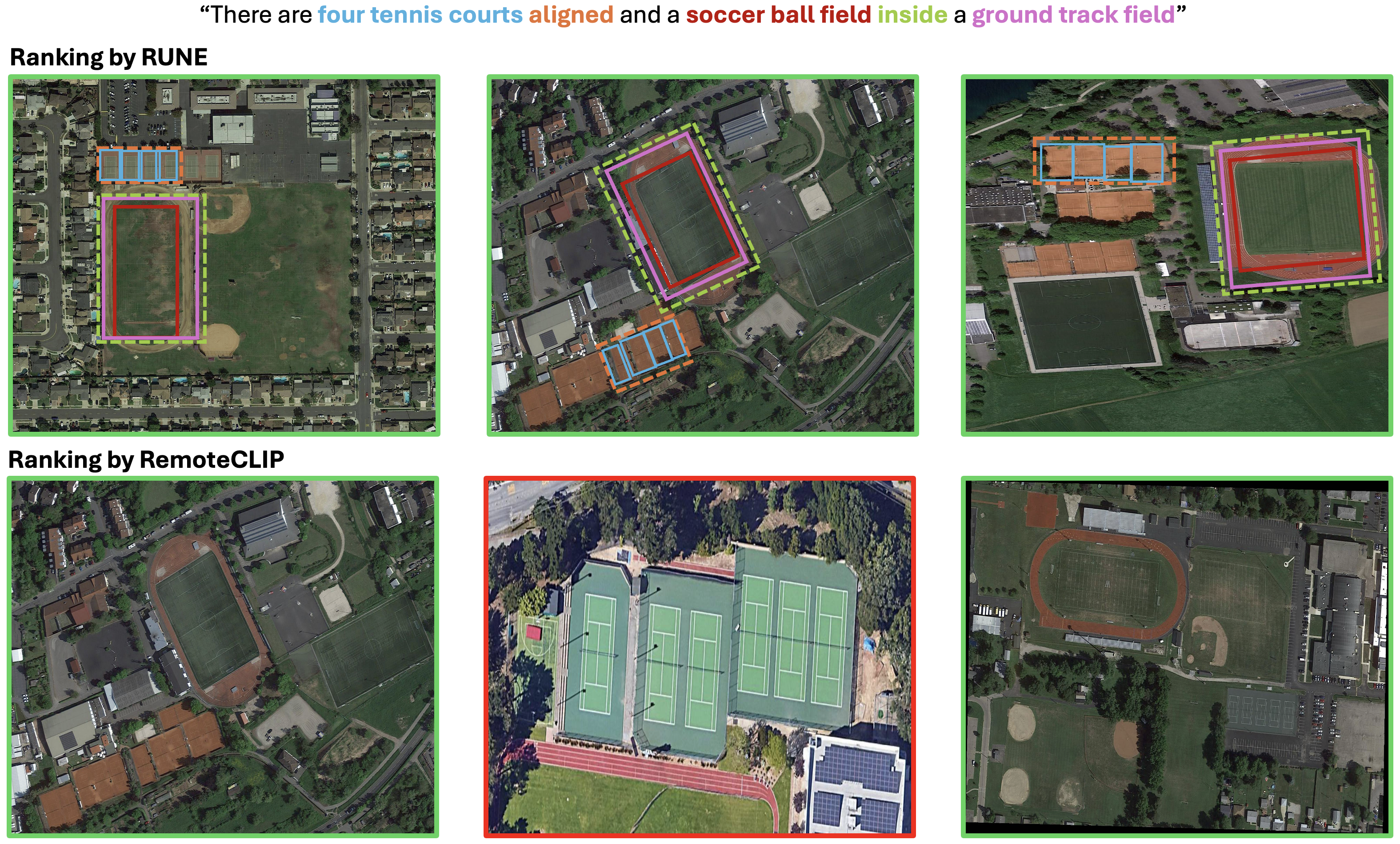}
         \caption{Retrieval of images containing tennis courts horizontally aligned and a soccer ball field inside a ground track field. While RUNE retrieves all relevant objects, RemoteCLIP wrongly retrieves the second object, as it does not contain any soccer ball field.}
         \label{fig:explainability1}
     \end{subfigure}
     \par\vspace{1em}
     \begin{subfigure}{\textwidth}
         \centering
         \includegraphics[width=0.90\textwidth]{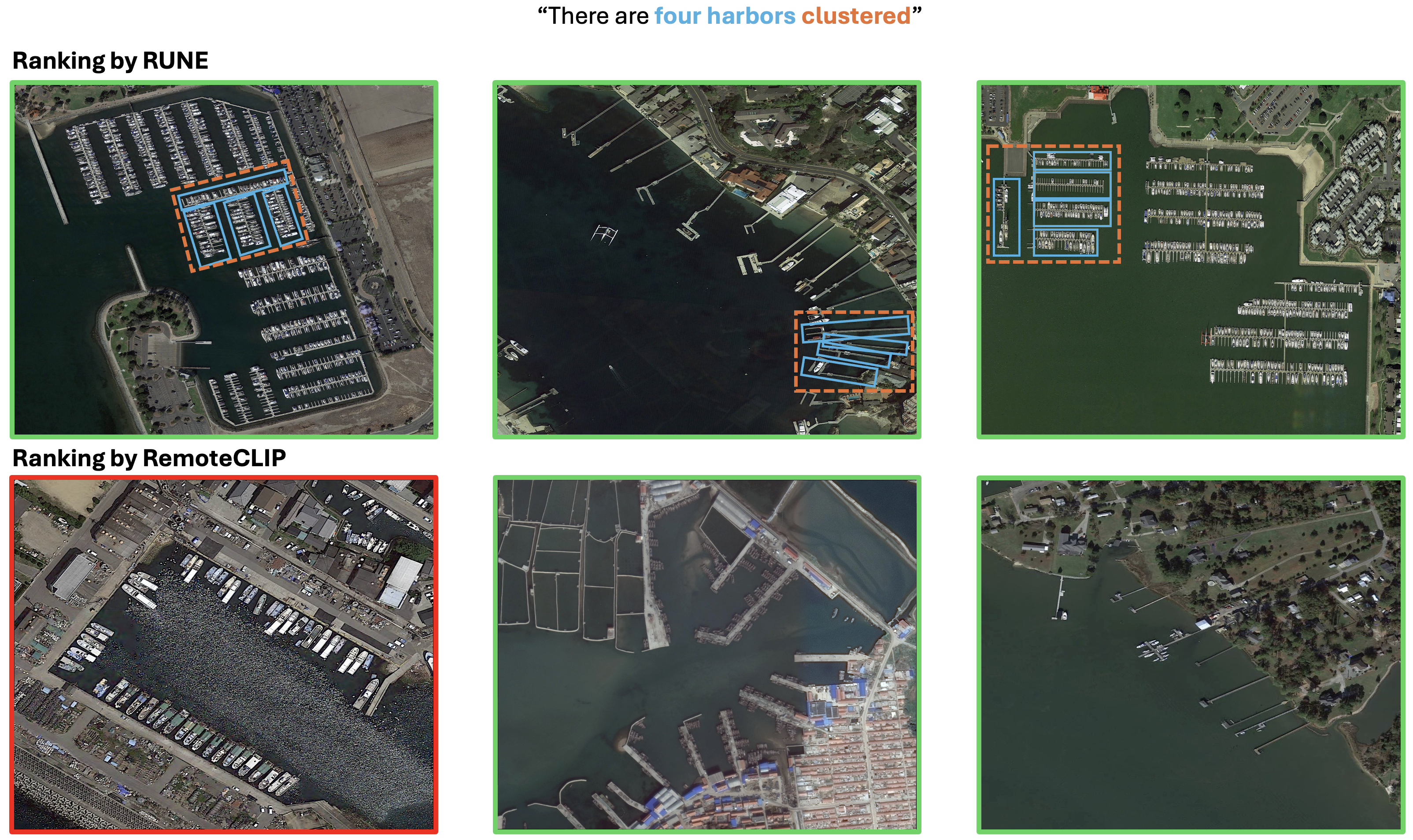}
         \caption{RUNE and RemoteCLIP retrieve images related to harbors cluster, thus posed in a shape that can resemble objects closely grouped together. RemoteCLIP retrieves the first object wrongly, as there are no harbors in the image.}
         \label{fig:explainability2}
     \end{subfigure}
    \caption{Visualization of retrieval results for RUNE and RemoteCLIP. Images are displayed from left to right, ordered from highest to lowest rank. RUNE, other than retrieving more relevant results leading to higher precision and recall, also guarantees greater explainability as allows to visualize the objects that were relevant in the retrieval of the results.}
    \label{fig:explainability}
\end{figure}

Figure \ref{fig:explainability} presents retrieval results for the compositional query: \emph{There are four tennis courts aligned and a soccer ball field inside a ground track field} (Figure \ref{fig:explainability1}) and \emph{There are four harbors clustered} (Figure \ref{fig:explainability2}). The figure contrasts the outputs of RUNE (top row) with those of RemoteCLIP (bottom row), the two leading models in terms of R@k and P@k performance.

RemoteCLIP computes image-text similarity using global embeddings. Thus, this approach lacks the ability to explicitly disentangle and ground compositional semantics, as it cannot isolate individual objects or enforce spatial and logical relations expressed in structured queries. This has two consequences which are both illustrated in Figure \ref{fig:explainability}. First, RemoteCLIP tends to retrieve also images that do not reflect the complexity of the input query in all its details. In the first example, it retrieves an image with four tennis courts but without a soccer ball field, and in the second, an image where there is a big harbor but lacking the specified cluster of four harbors. Second, explainability is significantly reduced, as even when the retrieved images are relevant, the link between the linguistic elements of the query and the visual content is less interpretable and verifiable.

RemoteCLIP computes image-text similarity using global embeddings. Thus, this approach lacks the ability to explicitly disentangle and ground compositional semantics, as it cannot isolate individual objects or enforce spatial and logical relations expressed in structured queries. This limitation leads to two main consequences, both illustrated in Figure \ref{fig:explainability}. First, RemoteCLIP retrieves images that fail to fully capture the complexity of the input query. For example, it returns an image with four tennis courts but no soccer ball field in the first case, and an image featuring a large harbor but missing the specified cluster of four harbors in the second case. Second, explainability is significantly reduced as even when the retrieved images are relevant, the connection between the linguistic components of the query and the visual content is less interpretable and harder to verify.

In contrast, RUNE first detects OBBs for candidate objects, then parses the query compositionally to map entities and spatial relations (e.g., “aligned,” “inside,” “clustered”) into logical forms. The logical forms are matched against the image by reasoning over object labels and spatial configurations. This approach offers a key advantage: \textbf{semantic grounding is preserved throughout the retrieval process.}

Each entity and relation in the query is explicitly mapped to visual evidence in the image, maintaining a traceable correspondence from natural language to visual structure. For instance, in the first example shown, all images retrieved by RUNE clearly display: (i) four tennis courts arranged in a linear alignment, and (ii) a soccer field situated within a larger elliptical ground track—precisely as dictated by the query. The same holds for the second example displayed: (i) four harbors closely grouped together. The corresponding OBBs highlight these regions and serve as visual proofs that the compositional constraints have been satisfied.

This explainability is intrinsic to RUNE’s design. By decomposing the query into interpretable components and matching them against symbolic representations of image content, RUNE allows human analysts to inspect whether the retrieval is both semantically accurate and logically coherent. The retrieval process is no longer a black box but an interpretable, compositional pipeline where intermediate reasoning steps are both visible and verifiable. In summary, RUNE achieves state-of-the-art performance while offering fine-grained semantic alignment and transparent interpretability.

\manu{
\subsection{Efficiency: VLMs vs RUNE}
\label{sec:efficiency}
\begin{figure}[h]
    \centering   \includegraphics[width=\linewidth]{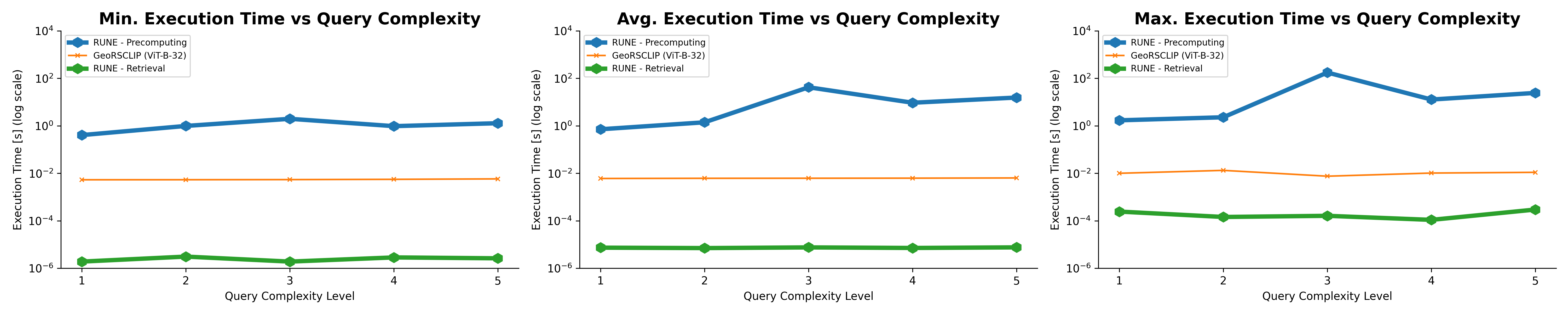}
    \caption{Comparison of execution time trends between the most efficient neural approach (GeoRSCLIP) and RUNE across increasing query complexity. RUNE is evaluated in two settings: RUNE – Precomputing, which includes predicate computation time, and RUNE – Retrieval, which measures only the time for matching predicates with bounding boxes.}
    \label{fig:efficiency}
\end{figure}

In this section, we compare the computational efficiency of baseline VLMs with RUNE. The results are illustrated in Figure \ref{fig:efficiency}, which shows the minimum, average, and maximum execution times for each query complexity level. We analyze these metrics because increasing query complexity leads to more entities and more complex relationships between objects.

Figure \ref{fig:efficiency} presents two computational scenarios. In the first scenario, the execution time of RUNE includes the time required to compute the predicates, which is an operation executed only once for an image. For retrieval, the predicates are then combined with the bounding boxes from the object detector to evaluate query-image compatibility. In this case, the most efficient VLM baseline (GeoRSCLIP) outperforms RUNE across minimum, average, and maximum execution times. RUNE’s execution time is generally on the order of seconds, with peaks reaching minutes, particularly for complexity level three (maximum execution time $\mathtt{\sim}10^{2}$ seconds).

In the second scenario, we consider only the compatibility calculation, simulating the case in which predicates are precomputed by the analyst. Under this setting, RUNE’s execution time is comparable to, and in some cases lower than, that of the baseline approaches. For example, the maximum execution time for RUNE is on the order of $\mathtt{\sim}10^{-4}$ seconds, which is lower than GeoRSCLIP’s maximum execution time of $\mathtt{\sim}10^{-2}$ seconds.

}

\manu{
\subsection{Flooding Scenario}
\label{subsec:flooding_scenario}

\begin{figure}[h]
    \centering
    \includegraphics[width=\linewidth]{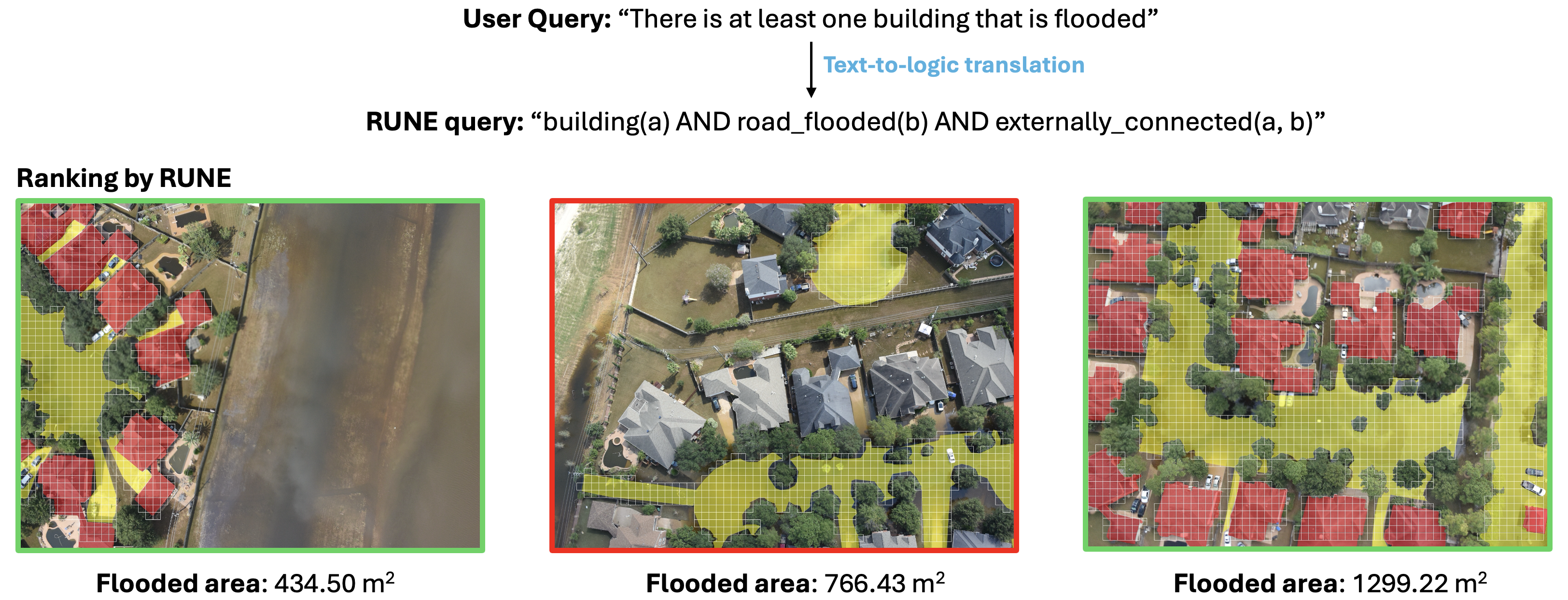}
    \caption{Retrieval of post-flood satellite images using RUNE. Images depict houses surrounded by flooded roads, with areas corresponding to houses and streets split into bounding boxes. These bounding boxes are used by RUNE to perform reasoning and retrieve relevant images. Performance metrics for this use case are as follows: R@1 = 0.10, P@1 = 1.00; R@5 = 0.40, P@5 = 0.80; R@10 = 0.90, P@10 = 0.90.}
    \label{fig:flooded_image}
\end{figure}

In this section, we demonstrate the functionality of RUNE for retrieving images depicting flooded houses. This use case is based on a subset of the FloodNet dataset \cite{rahnemoonfar2021floodnet}, which contains high-resolution post-flood scenes. Each image has dimensions of 4000 × 3000 pixels and was captured using a quadcopter flying at 200 feet (60.96 m) above ground level, equipped with a DJI FC220 camera with a 5 mm focal length and sensor dimensions of 6.16 mm × 4.55 mm \cite{stateczny2019shore}. To evaluate the ability of our methodology to prioritize images containing flooded buildings, we created a balanced test dataset comprising ten images with flooded buildings, which may also contain non-flooded buildings, and ten images containing only non-flooded buildings. 

As object detection techniques are insufficient for directly identifying flooded roads and relevant areas in post-flood scenarios, we rely on segmentation \cite{minaee2021image} to identify distinct regions within an image. To enable neurosymbolic reasoning, the segmented areas are converted into bounding boxes for houses and flooded roads, as illustrated by the white grids in Figure \ref{fig:flooded_image}. This transformation allows logic-based rules to be applied effectively. Since our primary goal is to demonstrate the application of neurosymbolic reasoning on segmented regions, we use the ground truth segmentations.

Figure \ref{fig:flooded_image} shows the user query and the first three retrieved results. The query requests all images containing at least one flooded building. While this natural language query can be directly used with LVLMs, it does not explicitly contain logic-expressible rules for the neurosymbolic engine. Following the methodology described in Section \ref{subsec:first_order_logic_rs}, the query is internally transformed into the following FOL expression:

\begin{align}
    building(a) \: AND \: road\_flooded(b) \: AND \: externally\_connected(a, b)
\end{align}

In this formulation, we model the flooding scenario using the relation externally connected (EC), that is a spatial relation defined by the RCC framework \cite{li2003region}, indicating that two regions share a boundary but their interiors do not overlap. This relation is appropriate in flood scenarios, where flooded buildings are typically surrounded by water; consequently, the bounding boxes for buildings are likely to share edges with those representing flooded roads, satisfying the externally connected condition.

Figure \ref{fig:flooded_image} also displays the flooded area, measured in square meters, for the retrieved images, illustrating how metadata from the acquisition process can be used to extract additional insights about the scenario. Specifically, to compute the flooded area, we leverage camera parameters—namely the focal length, sensor dimensions, and flight altitude—to calculate the ground sampling distance (GSD), which defines the real-world distance represented by each pixel in the image. The GSD for the image width and height is given by the following equations:

\begin{align}
    \label{eq:gsdw}
    GSD_w = \frac{\text{flight altitude} \cdot \text{sensor width}}{\text{focal length} \cdot \text{image width}} \\
    \label{eq:gsdh}
    GSD_h = \frac{\text{flight altitude} \cdot \text{sensor height}}{\text{focal length} \cdot \text{image height}}
\end{align}

where $GSD_w$ and $GSD_h$ denote the ground sampling distance along the width and height dimensions, respectively. In our setup, the computed GSD values are:

\begin{align}
    \label{eq:gsdw}
    GSD_w = \frac{\text{60.96m} \cdot \text{6.16mm}}{\text{5mm} \cdot \text{4000}} = 0.0188 \\
    \label{eq:gsdh}
    GSD_h = \frac{\text{60.96m} \cdot \text{4.55mm}}{\text{5mm} \cdot \text{3000}} = 0.0185
\end{align}

This implies that the width of each pixel correspond to 0.0188 meters, while the height correspond to 0.0185 meters. For each bounding box annotating the flooded regions, we can calculate the width in meters and the height in meters, by multiplying the width in pixel and the height in pixel respectively by the $GSD_w$ and the $GSD_h$. Then by multiplying the width and the height we derive the area in square meters and by summing all the bounding boxes areas we estimate the total flooded area in square meters.}

\section{Discussion and Limitations}
\label{sec:discussion}

The results support the hypothesis that the conjunction between foundation models and neurosymbolic AI can improve the performance of RS text-image retrieval. The improvement can be explained by the explicit modeling of spatial relations in images, which neurosymbolic AI can handle, while this is lost when images and text are transformed into embeddings and their similarity is measured. Besides performance and contrary to joint embedding approaches, RUNE also allows for explainability. RUNE indicates the involved objects and their relations on which the predicted probability was based. Our evaluation shows that LVLMs, which are usually evaluated only in terms of recall (R@k), have low performance in terms of precision (P@k), indicating that they tend to retrieve false positives and thus non-relevant images. This aspect is particularly critical, as it limits the possibility of employing LVLMs in real-life scenarios, where it is not possible to react to every image that the model flags. RUNE shows promising results in terms of precision, actually being able to improve the overall performance when increasing the complexity of the queries. This can be explained by considering neurosymbolic inference, as it allows for a more granular decomposition of the query without any loss of information given regarding its objects and relations.

Our method also has limitations. \manu{The repurposed DOTA dataset consists exclusively of cropped images. A possible extension of this work could focus on applying neurosymbolic reasoning to stitched and larger image frames, enabling the modeling and computation of more complex spatial relations.} \manu{The repurposed DOTA dataset is not aligned to a global or local coordinate system; therefore, queries cannot be formulated using cardinal directions (North, South, East, West). A possible extension of this work involves conducting experiments with datasets that include such coordinate mappings, thereby enabling the formulation of richer directional queries.}

While the text-to-logic transformation performed by the LLM is generally effective, it is not flawless. Errors in this step can propagate through the neurosymbolic reasoning pipeline, potentially leading to incorrect results. \manu{To improve the robustness and semantic clarity of the system, we plan to align our internal ontology with formal, standardized ontologies from the remote sensing and geospatial domains. This would also enhance interoperability and facilitate better generalization across different datasets and applications.} 

Other limitations include computational complexity and thus the execution time. Although the proposed conditional computation by logic decomposition reduced the computational complexity challenges are still present. Sometimes a logical clause can still have seven entities for instance when seven ships are arranged in a column. With many ship detections, this will still lead to a huge number of hypotheses, even with the proposed strategy. Possible solutions for reducing computational complexity include parallelizing the reasoning process across different groups of objects. 

\manu{Regarding the execution time, in case in which the predicates are precomputed, execution time is reduced at the level of neural approaches. However, the precomputation is only possible when the analyst knows in advance what must be found in the images and thus can use part of the analysis time to implement the precomputation. However, this may not always be possible as there can be cases in which the entirety of the analysis must be executed in real time, included the formulation of the query and thus the computation of the predicates.} 

Furthermore, the overall performance of the pipeline is highly dependent on the accuracy of the object detector. If the detector fails to identify the objects of interest in relevant images or misclassifies them, it can significantly impact the subsequent logic inference. A partial solution to this limitation is to utilize state-of-the-art (SOTA) object detectors that are specifically optimized and well-validated for the particular types of objects being targeted.

\section{Conclusions}
\label{sec:conclusions}
As the automation of RS tasks is becoming more widespread, it is fundamental to employ systems which are trustworthy and explainable. We proposed a new method RUNE, which fuses the capacities of foundational models and neurosymbolic AI to perform RS text-image retrieval by logical reasoning. Foundation models served as a text-to-logic converter and entity extraction, whereas probabilistic, logical reasoning served as an inference engine. We compared RUNE with three general-purpose LVLMs and three RS-LVLMs. We tested all models on the DOTA dataset repurposed for the task of text-image retrieval, and measured the performance of LVLMs for increasing query complexity using two retrieval robustness metrics. \manu{The results confirmed that RUNE outperforms the LVLMs in terms of performance, explainability, and robustness to query complexity}, as LVLMs struggled with more complex queries and showed low precision due to the inclusion of irrelevant images during the retrieval. \manu{Regarding robustness to image uncertainty, RUNE exhibited lower RRIU than four baselines due to its dependence on object detection; however, showing higher performance than that of most LVLMs evaluated.}
\manu{In the end, RUNE demonstrates efficient execution times when predicates are pre-calculated, as well as high effectiveness in retrieving post flood satellite imagery, highlighting its potential for deployment in real-world disaster-related scenarios.}

\section*{Acknowledgments}
We thank Fieke Hillerström for the insightful guidance and suggestions regarding the implementation of the symbolic engine. This work was partially supported by the \textit{Nederlandse Organisatie voor Wetenschappelijk Onderzoek (NWO)} under the KIC HEWSTI Project under grant no. KIC1.VE01.20.004. 

\section*{CRediT author statement}
Conceptualization: EM, GB; Methodology: EM, GB; Software: EM, GB, MK; Validation: EM; Formal analysis: na; Investigation: EM, GB; Resources: MK; Data Curation: EM; Writing - Original Draft: EM, GB; Writing - Review \& Editing: GB; Visualization: EM; Supervision: GB, MK; Project administration: GB, MK; Funding acquisition: MK;

\bibliographystyle{unsrt}  

\end{document}